\documentclass{article}
\usepackage{iclr2027_conference,times}
\iclrfinalcopy  
\usepackage[utf8]{inputenc}
\usepackage[T1]{fontenc}
\usepackage{hyperref,xurl,booktabs,amsfonts,nicefrac,microtype,xcolor}
\usepackage{graphicx,subcaption,amsmath,amssymb,bm,cleveref}
\crefname{appendix}{Appendix}{Appendices}
\Crefname{appendix}{Appendix}{Appendices}
\title{Screening Is Enough}
\author{%
Ken M. Nakanishi\\
Center for Emergent Matter Science (CEMS), RIKEN\\
Graduate School of Science, The University of Tokyo\\
\texttt{ken.m.nakanishi@gmail.com}
}
\begin{document}
\maketitle
\lhead{}  

\begin{abstract}
We call query--key relevance \emph{absolute} when its values lie on a fixed bounded scale, depend on neither competing keys nor sequence length, require no sequence-length-dependent calibration, and can all be zero.
To realize this notion, we introduce \emph{screening}, whose explicit threshold transforms bounded query--key similarities into relevance values, enabling exact rejection, empty selection, and direct inspection on a common scale.
In a controlled comparison of 12 attention mechanisms on a matched Transformer backbone, only screening maintains both low long-context perplexity and robust retrieval beyond the training context; notably, it does so without inference-time scaling.
Building on screening, we introduce Multiscreen, a language-model architecture composed of parallel gated screening tiles.
Multiscreen retains these long-context gains while achieving greater parameter efficiency, stronger general zero-shot downstream performance, lower training cost at larger scales, and lower model-side time to first token than Transformer baselines.
We further develop a normalization design that keeps Multiscreen training stable even at a learning rate of $1$ and show that an adapted version likewise stabilizes Transformer at the same learning rate.
\end{abstract}


\section{Introduction}

Handling contexts substantially longer than those seen during training remains a central challenge for large language models (LLMs).
Training directly on long sequences is computationally expensive because the cost of Transformer self-attention grows quadratically with sequence length~\citep{vaswani2017attention}.
Models are therefore often trained on relatively short contexts and expected to generalize to much longer ones at inference time, yet increasing the nominal context length does not guarantee that they can effectively use relevant information within it.

A key difficulty lies in how relevant information is selected from the context.
In standard softmax attention, scores are unbounded and scale-dependent, while weights redistribute a fixed unit mass, so a key's apparent relevance depends on competing keys.
Consequently, softmax cannot directly reject every key, and this competition can dilute relevant signals as contexts grow.

We call query--key relevance \emph{absolute} when its values lie on a fixed bounded scale, depend on neither competing keys nor sequence length, require no sequence-length-dependent calibration, and can all be zero.
To realize this notion, we introduce \emph{screening}.
Screening maps unit-normalized query--key similarities to nonnegative relevance values and uses an explicit threshold to assign exactly zero relevance to irrelevant keys.
Because these values are not normalized across keys, every key can be rejected, making both selected keys and empty selections directly inspectable on a common scale.

Validation loss alone does not reveal whether a model can retrieve and use relevant information from long contexts.
To directly evaluate retrieval, we introduce \textbf{ABCDigits}, a semantics-free key--value retrieval benchmark designed to isolate retrieval behavior from natural-language semantics and instruction-following effects.

We first isolate screening in a controlled comparison that changes only the attention mechanism on a matched Transformer backbone.
Among 12 evaluated variants, only screening maintains both low long-context perplexity and robust retrieval beyond the training context; notably, it requires no inference-time scaling.
Nested ablations isolate the contributions of screening's windowing and positional components, while showing that RoPE alone does not explain the abrupt failure of other global mechanisms beyond the training context.

Building on screening, we introduce \textbf{Multiscreen}, whose layers comprise parallel gated screening tiles with learned screening windows and minimal positional encoding (MiPE).
MiPE becomes inactive for sufficiently large windows, so broad-context tiles do not extrapolate positional rotations beyond those seen during training.

Multiscreen models retain these long-context gains: they maintain low perplexity without inference-time scaling throughout evaluations extending to $32\times$ the training context and show little retrieval degradation as context length increases.
In contrast, YaRN delays Transformer degradation but introduces a tradeoff between long-context extrapolation and perplexity at shorter contexts.

Beyond long-context behavior, the fitted scaling trends indicate that Multiscreen reaches similar validation loss with roughly 30\% fewer parameters, and the 1.3B Multiscreen model improves general zero-shot downstream performance, outperforming its Transformer counterpart on eight of nine tasks.
It also has lower training cost at larger scales and lower model-side time to first token (TTFT) across all evaluated model sizes.
At the 1.3B scale, the Transformer has $2.5$--$2.9\times$ the TTFT of Multiscreen across evaluated context lengths.

We further develop a normalization design that keeps Multiscreen training stable even at learning rate $1$.
Adapting this design to Transformer by combining query--key and pre-logit normalization likewise prevents divergence at the same learning rate, showing that this stability is transferable rather than specific to screening.

Our main contributions are as follows:
\begin{itemize}
\item We introduce \emph{screening}, a mechanism for absolute query--key relevance, and show through a controlled comparison of 12 attention variants that it alone maintains low long-context perplexity and robust retrieval without inference-time scaling.
\item We introduce \textbf{ABCDigits}, a semantics-free completion-based retrieval benchmark designed to isolate retrieval behavior without relying on natural-language semantics or instruction-following.
\item We build \textbf{Multiscreen} from parallel gated screening tiles and show that it retains screening's long-context gains while improving parameter efficiency, general zero-shot downstream performance, training cost, and model-side TTFT relative to Transformer baselines.
\item We develop a transferable normalization design that enables stable Multiscreen and Transformer training even at learning rate $1$.
\end{itemize}


\section{Related Work}

\paragraph{Attention mechanisms and relevance.}
Alternatives to standard softmax modify its normalization, sparsity, or accessible key set.
Normalized variants in our controlled comparison include Softmax$_1$, which adds a null alternative to the denominator~\citep{miller2023attention}; Scalable-Softmax (SSMax), which sharpens attention as sequence length grows~\citep{nakanishi2025scalable}; Entmax, which produces sparse normalized weights~\citep{peters2019sparse,correia2019adaptively,vasylenko2025long}; and Forgetting Attention, which applies data-dependent decay before normalization~\citep{lin2025forgetting}.
Sliding-window attention instead restricts the accessible keys~\citep{beltagy2020longformer}.
Sigmoid attention and $\mathrm{ReLU}^2$ attention depart from normalized competition by weighting query--key pairs independently~\citep{ramapuram2025theory,hua2022transformer}.
These mechanisms relax different constraints on total mass, sparsity, boundedness, pairwise independence, and length dependence; among those evaluated, screening alone combines all properties required for absolute relevance across context lengths.
The formal distinctions and complete evaluated configurations are summarized in \cref{tab:attention-properties,tab:attention-comparison}.

\paragraph{Long-context positional representations.}
Length generalization is also shaped by positional representations, including ALiBi, RoPE, and NoPE~\citep{press2022train,su2024roformer,kazemnejad2023impact}, as well as RoPE extension methods such as Position Interpolation, YaRN, and LongRoPE~\citep{chen2023extending,peng2024yarn,ding2024longrope}.
Our controlled comparison includes ALiBi, NoPE as used by Forgetting Attention, and RoPE with YaRN scaling.
Multiscreen instead uses minimal positional encoding (MiPE), which rotates only two dimensions for small screening windows and becomes inactive for sufficiently large windows.

\paragraph{Retrieval evaluation and synthetic benchmarks.}
Long-context evaluation has shown that models may fail to use relevant information even when it lies within the nominal context window, as in lost-in-the-middle phenomena~\citep{liu2024lost}.
Needle-in-a-haystack and passkey-style benchmarks probe long-context retrieval~\citep{kamradt2023niah,mohtashami2023random}.
For natural-language needle benchmarks, semantic interference from the surrounding context can affect retrieval results~\citep{shi2025semantic}.
The original passkey evaluation uses a fixed natural-language prompt template, and subsequent work shows that apparent retrieval performance can depend strongly on its prompt and passkey configuration~\citep{yuan2024kv}.
ABCDigits complements these benchmarks with a semantics-free completion task, a fixed key set, and a uniquely determined target value.


\section{Model Architecture}\label{sec:model}

\begin{figure}[tb]
\centering
\includegraphics[width=0.95\linewidth]{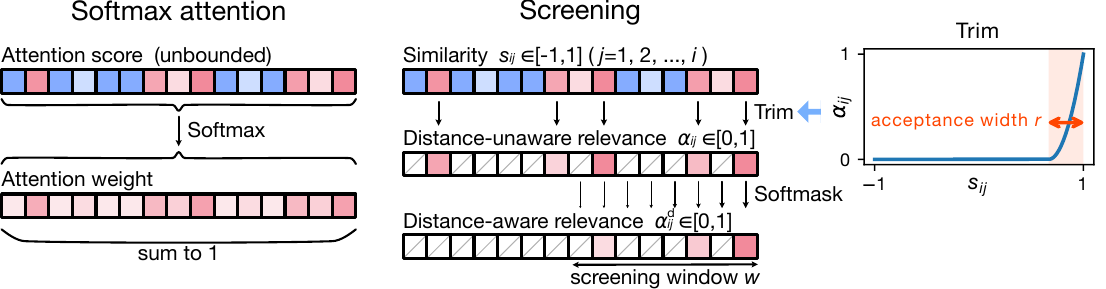}
\caption{
Conceptual comparison between softmax attention and screening.
Each cell corresponds to a scalar value associated with an individual key.
Colors indicate scalar values (blue: negative, red: positive; intensity indicates magnitude), while hatched cells denote zero values.
Softmax attention maps unbounded attention scores to nonnegative attention weights that sum to one, redistributing mass across all keys.
In contrast, screening maps bounded query--key similarities to distance-unaware relevance via Trim, which assigns exactly zero relevance to similarities below a threshold, and then applies Softmask to produce distance-aware relevance.
}
\label{fig:scrn-core}
\end{figure}

Multiscreen realizes absolute query--key relevance through \emph{screening}.
Screening begins with unit-normalized query--key dot products, yielding similarities on the fixed interval $[-1,1]$.
Trim and Softmask transform each similarity independently into a relevance value, without normalization across keys or calibration to the sequence length, and these values are used directly for value aggregation.
This construction permits exact rejection and empty selection, in contrast to the relative weighting imposed by softmax attention, as illustrated in \cref{fig:scrn-core}.

\begin{figure}[t]
\centering
\begin{subfigure}[c][49mm]{0.31\linewidth}
\vfill
\centering
\includegraphics[width=47mm]{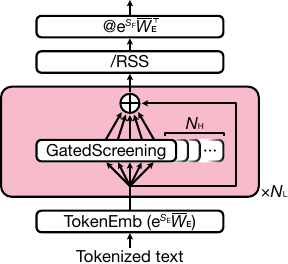}
\vfill
\caption{Multiscreen architecture}
\label{fig:musc}
\end{subfigure}
\hfill
\begin{subfigure}[c][49mm]{0.35\linewidth}
\vfill
\centering
\includegraphics[width=43mm]{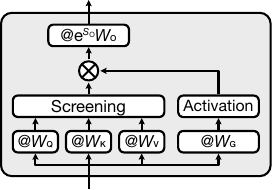}
\vfill
\caption{Gated screening tile}
\label{fig:gscrn}
\end{subfigure}
\hfill
\begin{subfigure}[c][49mm]{0.24\linewidth}
\vfill
\centering
\includegraphics[width=31mm]{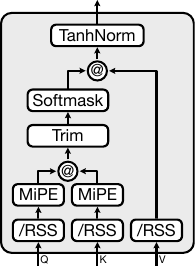}
\vfill
\caption{Screening unit}
\label{fig:scrn}
\end{subfigure}
\caption{
(a) Multiscreen comprises $N_\text{L}$ residual layers with $N_\text{H}$ parallel gated screening tiles per layer; $\overline{W}_\text{\!E}$ is the row-wise unit-normalized embedding matrix shared by the input and language-modeling head, and $\mathrm{e}^{s_\text{E}}$ and $\mathrm{e}^{s_\text{F}}$ are learned input and logit scales.
Final hidden states are unit-normalized before the head.
(b) Each gated screening tile applies query, key, value, and gate projections, screening, gating, and an output projection with learned residual-output scale $\mathrm{e}^{s_\text{O}}$.
(c) The screening unit applies QKV unit-length normalization, MiPE, Trim, Softmask, and TanhNorm.
In the diagrams, ``@'' denotes matrix multiplication and ``/RSS'' denotes row-wise unit-length normalization.
}
\label{fig:model-arch}
\end{figure}

The overall architecture is illustrated in \cref{fig:musc}.
Each layer contains parallel gated screening tiles (see \cref{fig:gscrn}), which project token representations into query, key, value, and gate vectors, apply a screening unit to retrieve relevant context, and gate the result using a GLU-style multiplicative mechanism~\citep{van2016conditional,dauphin2017language,shazeer2020glu} before projecting back to the embedding dimension.
In the actual implementation, several operations are fused, and query--key pairs outside finite learned screening windows are skipped for efficiency.

\subsection{Overview}

Let $W_\text{E} = [\bm{e}_1; \dots; \bm{e}_{|\mathcal{V}|}]$ denote the embedding matrix shared by the input and language-modeling head.
We apply \emph{embedding normalization} to each row, defining $\bar{\bm{e}}_j = \bm{e}_j / \|\bm{e}_j\|$.
Given a token $t_i$, its input representation is
\begin{equation}
\bm{x}_i^{(0)} = \mathrm{e}^{s_\text{E}} \bar{\bm{e}}_{t_i},
\end{equation}
where $s_\text{E}$ is a learned scalar.

Multiscreen then applies $N_\text{L}$ residual layers, each composed of $N_\text{H}$ parallel gated screening tiles.
Let $\Delta \bm{x}_i^{(\ell,h)}$ denote the update produced for position $i$ by tile $h$ in layer $\ell$.
The layer update is
\begin{equation}
\bm{x}_i^{(\ell)}
=
\bm{x}_i^{(\ell-1)}
+
\sum_{h=1}^{N_\text{H}}
\Delta \bm{x}_i^{(\ell,h)}.
\end{equation}

After the final layer, we apply \emph{pre-logit normalization} (PLNorm), $\bar{\bm{x}}_i^{(N_\text{L})}=\bm{x}_i^{(N_\text{L})}/\|\bm{x}_i^{(N_\text{L})}\|$, and compute logits using the normalized hidden state and embeddings:
\begin{equation}
z_{ij}
=
\bar{\bm{x}}_i^{(N_\text{L})}
\left(
\mathrm{e}^{s_\text{F}} \bar{\bm{e}}_j^\top
\right),
\qquad
j \in \{1,\dots,|\mathcal{V}|\},
\end{equation}
where $s_\text{F}$ is a learned scalar.
Neither embedding normalization nor PLNorm includes learned elementwise scales; each only projects its input onto the unit hypersphere.
We refer to the combination of these normalizations and the learned input and logit scales $\mathrm{e}^{s_\text{E}}$ and $\mathrm{e}^{s_\text{F}}$ as the \emph{spherical input--output (SIO) parameterization}.
A standard softmax over logits gives next-token probabilities.

\subsection{Screening Unit}\label{sec:scrn}

A screening unit takes projected query, key, and value vectors $\bm{q}_i,\bm{k}_i \in \mathbb{R}^{1\times d_\text{K}}$ and $\bm{v}_i \in \mathbb{R}^{1\times d_\text{V}}$, and returns a context-dependent representation $\bm{u}_i \in \mathbb{R}^{1\times d_\text{V}}$ (see \cref{fig:scrn}).
Each unit has two learned scalar parameters, $s_\text{w}$ and $s_\text{r}$, which define
\begin{equation}
w = \mathrm{e}^{s_\text{w}} + 1,
\qquad
r = \sigma(s_\text{r}),
\end{equation}
where $\sigma$ denotes the sigmoid function.
Here $w$ is the screening window and $r$ is the acceptance width for similarity.
Under our \emph{default inference rule}, if the learned screening window exceeds the maximum sequence length seen during training, we set $w=\infty$.
All inference evaluations use this rule unless a finite-window ablation is explicitly identified.

We first apply unit-length normalization to queries, keys, and values:
\begin{equation}
\bar{\bm{q}}_i = \frac{\bm{q}_i}{\|\bm{q}_i\|},
\qquad
\bar{\bm{k}}_i = \frac{\bm{k}_i}{\|\bm{k}_i\|},
\qquad
\bar{\bm{v}}_i = \frac{\bm{v}_i}{\|\bm{v}_i\|}.
\end{equation}
In implementation, the denominator in each unit-length normalization is clipped from below by a small positive constant for numerical stability.
Normalizing queries and keys bounds query--key similarities in $[-1,1]$, making thresholding well-defined.
Prior work shows that token contributions depend on vector norms as well as attention weights~\citep{kobayashi2020attention,guo2024attention}.
We therefore normalize value vectors to unit length so that differences in their norms do not directly determine their contribution magnitudes.

\begin{figure}[tb]
\centering
\includegraphics[width=0.825\linewidth]{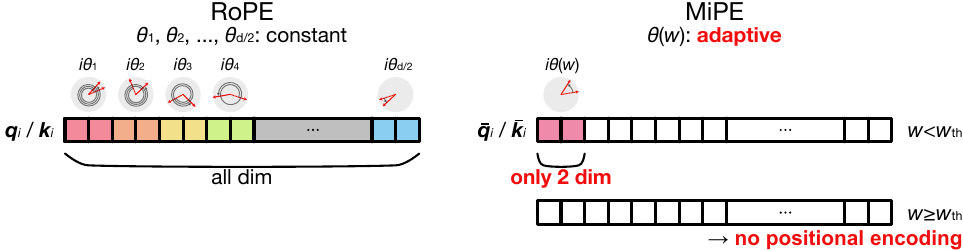}
\caption{
Comparison between RoPE and MiPE.
RoPE applies rotations across all dimensions with fixed frequencies.
MiPE applies rotation only to two dimensions and adapts the rotation angle based on the learned screening window $w$, becoming inactive when $w \ge w_\text{th}$.
}
\label{fig:mipe}
\end{figure}

Before computing similarity, we apply minimal positional encoding (MiPE; see \cref{fig:mipe}) to $\bar{\bm{q}}_i$ and $\bar{\bm{k}}_i$.
MiPE is a RoPE-like rotation~\citep{su2024roformer} applied only to the first two dimensions, with a rotation rate modulated by the learned screening window $w$.
For position $i$, MiPE rotates by angle $\phi(i,w)=i\theta(w)$, where
\begin{equation}
\theta(w)=
\begin{cases}
\frac{\pi}{w} \left(1-\frac{w}{w_\text{th}}\right),
& w < w_\text{th},\\
0,
& w \ge w_\text{th}.
\end{cases}
\end{equation}
Here $w_\text{th}$ is a fixed MiPE threshold.
Thus MiPE is active only for small screening windows and becomes the identity when $w \ge w_\text{th}$.

Let $\tilde{\bm{q}}_i$ and $\tilde{\bm{k}}_j$ denote the resulting position-encoded query and key vectors.
The bounded similarity is then
\begin{equation}
s_{ij} = \tilde{\bm{q}}_i \tilde{\bm{k}}_j^\top,
\qquad
s_{ij} \in [-1,1].
\end{equation}

We apply a Trim transform (see \cref{fig:scrn-core}) to obtain distance-unaware relevance:
\begin{equation}
\alpha_{ij}=
\left[
\max\left(1-\frac{1-s_{ij}}{r},0\right)
\right]^2.
\end{equation}
This sets relevance exactly to zero whenever $s_{ij} \le 1 - r$, and smoothly emphasizes similarities close to one.

We next apply a causal distance-aware Softmask parameterized by the screening window $w$:
\begin{equation}
m_{ij}(w)=
\begin{cases}
1-2\left(\frac{i-j}{w}\right)^2, & 0 \le i-j < \frac{w}{2},\\
2\left(1-\frac{i-j}{w}\right)^2, & \frac{w}{2} \le i-j < w,\\
0, & \text{otherwise}.
\end{cases}
\end{equation}
This piecewise-quadratic form provides a cosine-like taper while avoiding trigonometric evaluation.
For $w=\infty$, we take $m_{ij}(w)=1$ at every causal position.
We then define distance-aware relevance
\begin{equation}
\alpha^\text{d}_{ij}=\alpha_{ij}m_{ij}(w).
\end{equation}
A key contributes only if it survives both content-based and distance-based screening.

Finally, the screening unit aggregates the surviving values:
\begin{equation}
\bm{h}_i=
\sum_{j \le i}
\alpha^\text{d}_{ij}\bar{\bm{v}}_j.
\end{equation}
Because the relevance values $\{\alpha^\text{d}_{ij}\}_{j\le i}$ are not constrained to sum to one, the unit can represent the absence of relevant context.
To softly bound the output norm while preserving direction, we define the function
\begin{equation}
\operatorname{TanhNorm}(\bm{x})=
\frac{\tanh \|\bm{x}\|}{\|\bm{x}\|}\bm{x}.
\end{equation}
For sufficiently small $\|\bm{x}\|$, we use the limiting value $\operatorname{TanhNorm}(\bm{x})=\bm{x}$.
The screening output is
\begin{equation}
\bm{u}_i=
\operatorname{TanhNorm}(\bm{h}_i).
\end{equation}

Overall, screening defines relevance independently for each key, can assign exactly zero relevance to irrelevant keys, and aggregates only the surviving values.

\subsection{Gated Screening Tile}\label{sec:gscrn}

A gated screening tile (see \cref{fig:gscrn}) combines screening-based context retrieval with input-dependent feature selection.
Given token representations $\bm{x}_i \in \mathbb{R}^{1\times d_\text{E}}$, the tile computes
\begin{equation}
\bm{q}_i = \bm{x}_i W_\text{Q},
\quad
\bm{k}_i = \bm{x}_i W_\text{K},
\quad
\bm{v}_i = \bm{x}_i W_\text{V},
\quad
\bm{g}_i = \bm{x}_i W_\text{G},
\end{equation}
where $W_\text{Q},W_\text{K}\in\mathbb{R}^{d_\text{E}\times d_\text{K}}$ and $W_\text{V},W_\text{G}\in\mathbb{R}^{d_\text{E}\times d_\text{V}}$.

Using the screening unit from \cref{sec:scrn}, the tile retrieves
\begin{equation}
\bm{u}_i
=
\operatorname{Screening}
\bigl(
\bm{q}_i,\{\bm{k}_j\}_{j\le i},\{\bm{v}_j\}_{j\le i},w,r
\bigr)_i.
\end{equation}
In parallel, it computes a gate via a nonlinear activation
\begin{equation}
\hat{\bm{g}}_i
=
\tanh(\operatorname{SiLU}(\bm{g}_i)),
\end{equation}
where SiLU is the sigmoid-weighted linear unit~\citep{elfwing2018sigmoid}.
The tile output is
\begin{equation}
\Delta \bm{x}_i
=
\left(
\bm{u}_i \odot \hat{\bm{g}}_i
\right)
\left(
\mathrm{e}^{s_\text{O}} W_\text{O}
\right),
\end{equation}
with $W_\text{O}\in\mathbb{R}^{d_\text{V}\times d_\text{E}}$ and learned scalar $s_\text{O}$.

This tile can be viewed as a generalization of GLU-style modules~\citep{van2016conditional,dauphin2017language,shazeer2020glu} and recent gated-attention variants~\citep{qiu2025gated}, replacing the usual linear or attention-based transformation with screening-based context aggregation.
Thus, each tile jointly performs context selection, feature gating, and projection back to the model space.
Initialization details are provided in \cref{apx:init}.
Architectural analyses of MiPE, TanhNorm, and the output gate are provided in \cref{apx:mipe,apx:tanhnorm,apx:gate}.


\section{Experiments}

We compare Multiscreen with Transformer baselines across long-context language modeling, retrieval, absolute relevance, optimization stability, parameter efficiency, general zero-shot downstream performance, training cost, and model-side TTFT.

\subsection{Experimental Setup}\label{sec:setup}

Models in the main experiments are pretrained on SlimPajama~\citep{cerebras2023slimpajama} for $2^{38}$ tokens using sequences of length $2^{12}$.
Both architectures use AdamW~\citep{loshchilov2017decoupled} without weight decay or gradient clipping.
Transformer uses model-size-dependent learning rates, whereas Multiscreen uses the larger learning rate $2^{-4}$, enabled by its improved stability.
Full model configurations, training hyperparameters, and implementation details are provided in \cref{apx:setup}.

\subsection{Long-Context Perplexity}\label{sec:lcp}

\begin{figure}[tb]
\centering
\includegraphics[width=0.5\linewidth]{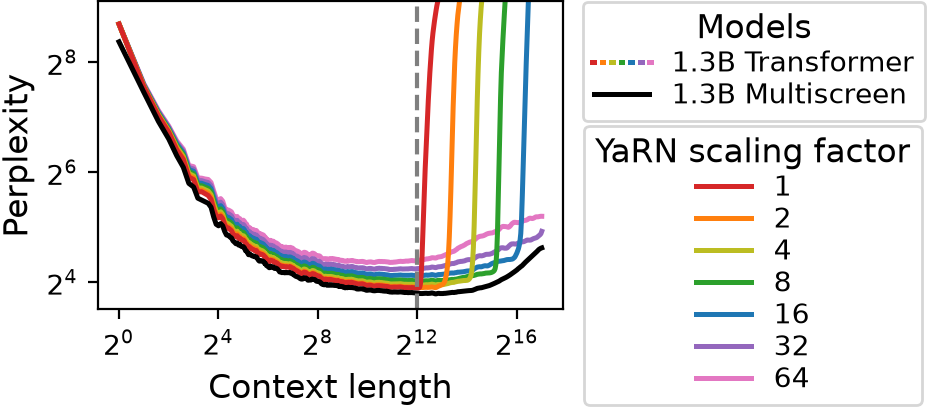}
\caption{
Position-dependent perplexity of the 1.3B models.
Colored curves show Transformer with different YaRN scaling factors, whereas the black curve shows Multiscreen, which requires no inference-time scaling.
The dashed line marks the pretraining context length ($2^{12}$).
}
\label{fig:lcp}
\end{figure}

We evaluate long-context language modeling using position-dependent perplexity.
As shown in \cref{fig:lcp}, Multiscreen requires no inference-time scaling and maintains low perplexity up to $32\times$ the training context length.
YaRN delays the onset of Transformer degradation, but larger scaling factors increase perplexity at shorter contexts, revealing a tradeoff that is absent from Multiscreen.
Additional evaluation details and seed-averaged smaller-model results are provided in \cref{apx:lcp}.

\subsection{Retrieval Ability}\label{sec:abcd}

\begin{figure}[tb]
\centering
\begin{subfigure}[c]{0.2\linewidth}
\centering
\begin{minipage}[c]{0.5\linewidth}
{\ttfamily
A=967892\\[-2.3pt]
W=900383\\[-2.3pt]
J=707723\\[-2.3pt]
[...]\\[-2.3pt]
W=900383\\[-2.3pt]
\textcolor[HTML]{cc0000}{L=169428}\\[-2.3pt]
A=967892\\[-2.3pt]
[...]\\[-2.3pt]
A=967892\\[-2.3pt]
\textcolor[HTML]{cc0000}{L=}
}
\end{minipage}
\caption{Prompt example}
\label{fig:prompt}
\end{subfigure}
\hfill
\begin{subfigure}[c]{0.775\linewidth}
\centering
\includegraphics[width=\linewidth]{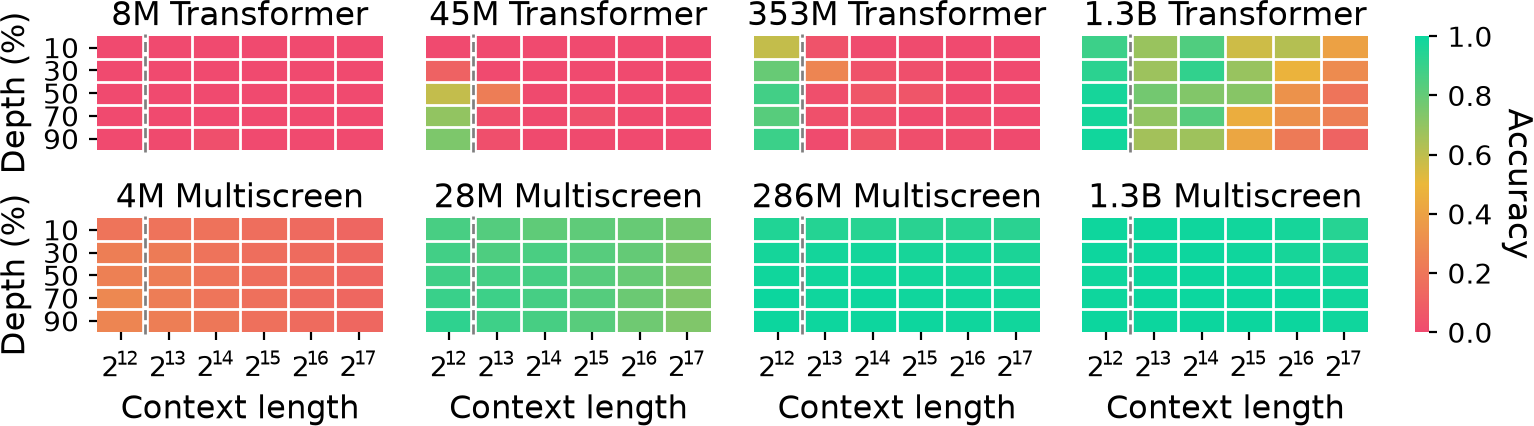}
\caption{Retrieval accuracy}
\label{fig:abcd}
\end{subfigure}
\caption{
(a) Example ABCDigits prompt.
(b) ABCDigits exact-match accuracy across model sizes, context lengths, and target depths.
Each cell averages 1,000 instances; results are further averaged across three independently trained models, except at 1.3B, where one model is used.
Transformer uses context-matched YaRN scaling, while Multiscreen uses our default inference rule.
The dashed line marks the pretraining context length ($2^{12}$).
}
\label{fig:prompt-abcd}
\end{figure}

To isolate retrieval ability, we introduce \textbf{ABCDigits}, a synthetic key--value retrieval benchmark in a semantics-free setting.
The task presents a shuffled list of mappings from uppercase letters to six-digit integers (e.g., \texttt{A=967892}), followed by a query (e.g., \texttt{L=}) requiring the model to retrieve the corresponding value.
The target mapping appears exactly once in the context, requiring the model to locate a unique occurrence.
The number of distinct keys is fixed to avoid key-cardinality confounds, while non-target frequencies are highly skewed so that the unique target does not stand out by rarity alone.
Code for generating ABCDigits prompts is publicly available.\footnote{\url{https://github.com/ken-nakanishi/abcdigits}}
Additional task-construction and evaluation details are provided in \cref{apx:abcd}.

We evaluate exact-match retrieval under greedy decoding at target context lengths from $2^{12}$ to $2^{17}$ and varying target depths.
As shown in \cref{fig:prompt-abcd}, Multiscreen shows little degradation with increasing context length and maintains high retrieval accuracy beyond the training context length.
Notably, even at the training context length, the 286M Multiscreen model outperforms the 1.3B Transformer evaluated with context-matched YaRN scaling when averaged over target depths (98.33\% vs. 95.18\%), with the gap becoming more pronounced at longer contexts.
In contrast, Transformer accuracy degrades increasingly beyond the training context despite using context-matched YaRN scaling.

With roughly one fifth the parameters, the 286M Multiscreen model achieves substantially stronger long-context retrieval than the 1.3B Transformer despite higher validation loss, showing that validation loss alone does not capture retrieval ability.
On the passkey retrieval task introduced by \citet{mohtashami2023random}, both the 286M and 1.3B Multiscreen models achieve perfect accuracy at every evaluated context length and target depth (\cref{apx:passkey}).

\subsection{Isolating the Screening Mechanism}\label{sec:screening-analysis}

To isolate screening from the remaining Multiscreen architecture, we train a controlled suite that matches the 353M Transformer in all model dimensions (including Q/K/V dimensions), initialization, and training hyperparameters, changing only the attention mechanism.
The alternatives include Softmax$_1$, SSMax, Entmax ($\alpha=1.5$), Sigmoid, $\mathrm{ReLU}^2$, ALiBi, Forgetting Attention, and sliding-window attention, spanning normalization-based, sparse, noncompetitive, distance-decayed, and fixed-window mechanisms.
The Screening variant contains the screening unit---QKV unit-length normalization, MiPE, Trim, Softmask, and TanhNorm---but excludes the output gate and the other components of the gated screening tile.
We include two nested ablations: Screening (RoPE) replaces MiPE with RoPE, while Screening (no window) additionally removes the learned window and Softmask.

\begin{figure}[t]
\centering
\includegraphics[width=\textwidth]{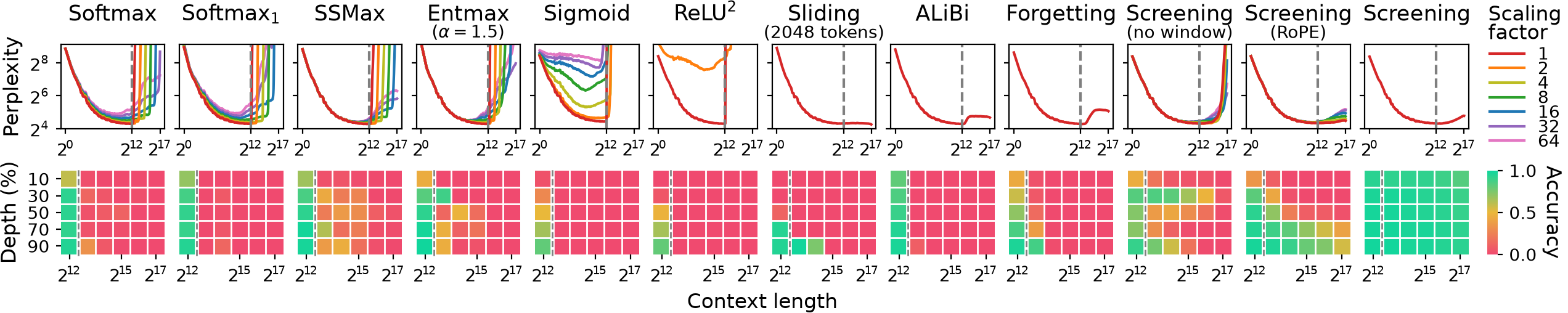}
\caption{
Controlled attention-mechanism comparison on the 353M Transformer backbone with all other settings fixed.
The top row shows position-dependent perplexity; the bottom row shows ABCDigits retrieval accuracy.
Each variant uses one trained model.
In the top row, red denotes no inference-time scaling; for methods requiring scaling, red corresponds to factor $1$ and the other colors to the indicated factors.
For ABCDigits, methods requiring scaling use a factor equal to the evaluation-to-training context-length ratio.
For $\mathrm{ReLU}^2$, scaling factors $4$ and above exceed the plotted perplexity range.
Dashed lines mark the pretraining context length ($2^{12}$).
}
\label{fig:attention-comparison}
\end{figure}

Screening is the only evaluated mechanism that maintains both low long-context perplexity and robust retrieval across all evaluated context lengths and target depths without inference-time scaling (\cref{fig:attention-comparison}).
ALiBi, Forgetting Attention, and sliding-window attention maintain relatively stable long-context perplexity yet fail to retrieve distant targets, showing that perplexity alone is insufficient to assess long-range retrieval.
Even at scaling factor $1$, Screening (no window) avoids the abrupt perplexity degradation beyond the training context seen in the other full-context RoPE-based mechanisms, showing that RoPE alone does not explain this failure.
Screening (RoPE) markedly outperforms Screening (no window), demonstrating the importance of the learned window and its associated Softmask.
Full Screening further improves retrieval by replacing RoPE with MiPE, indicating that MiPE provides additional robustness.
Together, these results isolate the long-context behavior to the screening unit itself rather than the output gate or the wider Multiscreen architecture.
Formal property comparisons, complete configurations, and implementation details are provided in \cref{apx:attention-comparison}.

\subsection{Learning-Rate Stability}\label{sec:lr}

We sweep the learning rate for a 45M Transformer and a 28M Multiscreen while keeping all other training settings fixed within each architecture.
For comparison, we also evaluate a Transformer control with weight decay $0.1$ and gradient clipping at $1.0$.

\begin{figure}[tb]
\centering
\hfill
\begin{subfigure}[c]{0.413\linewidth}
\centering
\includegraphics[width=\linewidth]{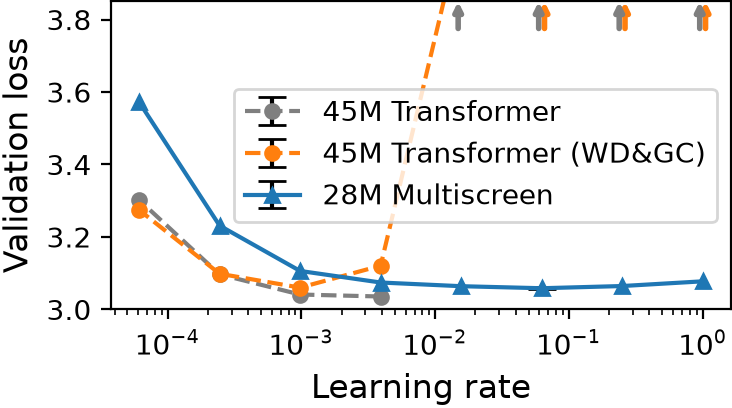}
\caption{Learning-rate sweep.}
\label{fig:lr-sweep-curve}
\end{subfigure}
\hfill
\begin{subfigure}[c]{0.5\linewidth}
\centering
\small
\begin{tabular}{@{}lc@{}}
\toprule
Configuration & Outcome \\
\midrule
Transformer & Diverged \\
Transformer $+$ SIO & Diverged \\
Transformer $+$ PLNorm & Diverged \\
Transformer $+$ QKNorm & Diverged \\
Transformer $+$ QKNorm $+$ SIO & Stable \\
Transformer $+$ QKNorm $+$ PLNorm & Stable \\
Multiscreen & Stable \\
\bottomrule
\end{tabular}
\caption{Selected outcomes at learning rate $1$.}
\label{fig:lr-components-selected}
\end{subfigure}
\hfill
\caption{
Learning-rate stability of Transformer and Multiscreen.
(a) Validation loss across learning rates, with arrows at the upper plot boundary indicating divergence.
Each point uses one training run, except Multiscreen at learning rate $2^{-4}$, which reuses the three runs from the scaling experiments and shows their mean and standard deviation.
(b) Training outcomes at learning rate $1$ after $2^{36}$ tokens ($2^{14}$ steps), where ``Stable'' indicates completion of this budget without divergence.
PLNorm denotes pre-logit normalization.
Complete results are provided in \cref{apx:lr-components}.
}
\label{fig:lr-sweep}
\end{figure}

As shown in \cref{fig:lr-sweep-curve}, both baseline Transformer configurations become unstable and eventually diverge beyond moderate learning rates, whereas Multiscreen remains stable at learning rate $1$.
This large-learning-rate stability transfers to Transformer when QKNorm is combined with either PLNorm or SIO (\cref{fig:lr-components-selected}).
At learning rate $1$, QKNorm, PLNorm, and SIO each fail to prevent Transformer divergence when applied alone.
These results show that large-learning-rate stability depends on how normalization components are combined and placed.
Additional training-loss and gradient-norm analyses are provided in \cref{apx:lr-log,apx:grad-norm}.
Complete Transformer and Multiscreen ablations are provided in \cref{apx:lr-components}.

\subsection{Scaling and Computational Efficiency}\label{sec:scaling}
\label{sec:efficiency}

\paragraph{Scaling and training cost.}
We evaluate scaling behavior by comparing models trained on a fixed budget of $2^{38}$ tokens.
For smaller models, we report mean and standard deviation over three runs with different random seeds; larger models are evaluated with a single run.
We report training cost in GPU hours, computed as elapsed wall-clock training time multiplied by the number of GPUs used.
All training runs use NVIDIA GH200 GPUs, with data-parallel configurations detailed in \cref{apx:setup}.

\begin{figure}[t]
\centering
\includegraphics[width=\linewidth]{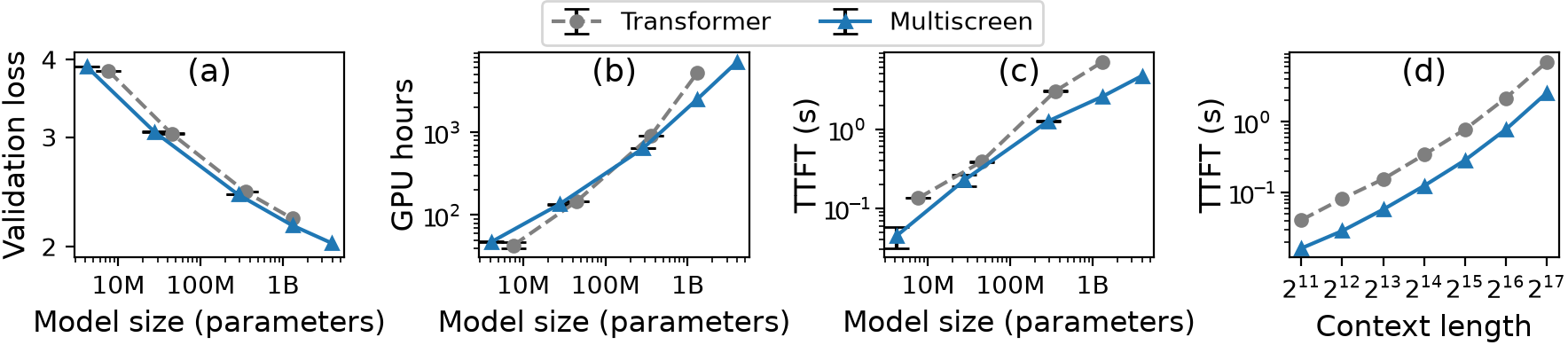}
\caption{
Scaling and computational efficiency.
(a) Validation loss and (b) training cost in GPU hours across model sizes, with each model trained on $2^{38}$ tokens.
(c) Model-side TTFT across model sizes at context length $2^{17}$.
(d) Model-side TTFT across context lengths for the 1.3B models.
Error bars indicate standard deviation across independently trained models where multiple runs are available.
}
\label{fig:scaling-training}
\end{figure}

As shown in \cref{fig:scaling-training}a, the fitted scaling trends indicate that Multiscreen requires roughly 30\% fewer parameters to reach a similar validation loss.
We evaluate general zero-shot downstream performance of the 1.3B models on nine common benchmarks.
Multiscreen outperforms Transformer on eight of the nine tasks and improves the unweighted mean accuracy from 52.41\% to 53.67\%, with per-task results in \cref{apx:downstream} and additional scaling analyses in \cref{apx:scaling}.
\Cref{fig:scaling-training}b further shows lower training cost at larger model scales, with a particularly pronounced advantage for the 1.3B models.

\paragraph{Model-side TTFT.}
We define model-side time to first token (TTFT) as the latency of a full-context forward pass that processes all positions through the model layers but applies the vocabulary projection only at the final position to produce first-token logits.
This metric isolates model execution and excludes tokenization, sampling, request queuing, and network or serving overhead.
We evaluate TTFT across model sizes at context length $2^{17}$ and across context lengths from $2^{11}$ to $2^{17}$ for the 1.3B models.
All measurements use a single NVIDIA GH200 GPU; the complete measurement protocol is provided in \cref{apx:ttft}.
As shown in \cref{fig:scaling-training}c, Multiscreen achieves consistently lower model-side TTFT than Transformer across model sizes.
The context-length sweep in \cref{fig:scaling-training}d likewise shows that Transformer has $2.5$--$2.9\times$ the TTFT of Multiscreen at every evaluated input length.


\section{Conclusion}

In this work, we introduced \emph{screening} to realize absolute query--key relevance on a fixed bounded scale, independently of competing keys and sequence length.
Its explicit threshold permits exact rejection and empty selection without sequence-length-dependent calibration, making relevance directly inspectable.

Using ABCDigits and long-context perplexity, a controlled comparison of 12 attention mechanisms showed that only screening maintained robust retrieval and low perplexity beyond the training context without inference-time scaling.
Multiscreen, built from parallel gated screening tiles, retained these long-context gains across all evaluated context lengths.

Beyond long-context behavior, Multiscreen improved parameter efficiency, general zero-shot downstream performance, training cost at larger scales, and model-side time to first token.
We also developed a normalization design that kept Multiscreen stable even at a learning rate of $1$ and whose adaptation likewise stabilized Transformer.

These results show that directly inspectable absolute relevance can coexist with strong long-context behavior, general zero-shot downstream performance, and computational efficiency, offering an alternative to competitive renormalization and length-dependent inference-time calibration.

\subsection*{Ethics Statement}

This work does not involve human participants or the collection of new personal data, and ABCDigits is constructed entirely from synthetic letter--digit mappings.
Our models are trained and evaluated on publicly released datasets whose provenance and licenses are documented in \cref{apx:licenses}.
Because SlimPajama contains web-derived data, it may reflect biases or harmful content and may contain personal or copyrighted information; we did not independently audit the complete corpus for these issues.

\subsection*{AI Use Statement}

Generative AI tools were used for literature retrieval; feedback on experimental design and result interpretation; implementation and debugging of selected baseline and diagnostic methods; and drafting, editing, and presentation of the manuscript.
The Multiscreen architecture and its core implementation were developed by the author without generative AI assistance.
All AI-assisted text, code, and scientific suggestions were reviewed by the author, experimental interpretations were checked against the reported results, and suggested references were verified against the original sources.
The author takes responsibility for the final content of this work.

\subsection*{Reproducibility Statement}

The Multiscreen architecture and screening mechanism are fully specified in \cref{sec:model}.
\Cref{apx:setup,apx:musc-config,apx:tf-config,apx:init} provide the training setup, model configurations, baseline details, and initialization schemes.
Additional scaling and downstream results are provided in \cref{apx:scaling,apx:downstream}.
Learning-rate analyses and detailed protocols for the long-context, retrieval, attention-mechanism comparison, and component-analysis experiments are provided in \cref{apx:lr-log,apx:grad-norm,apx:lr-components,apx:lcp,apx:abcd,apx:passkey,apx:attention-comparison,apx:mipe,apx:tanhnorm,apx:gate}.
The TTFT measurement protocol and the datasets and software used in this work are documented in \cref{apx:ttft,apx:licenses}.
Public code for generating ABCDigits prompts is linked in \cref{sec:abcd}.

\subsubsection*{Author Contributions}

Ken M. Nakanishi solely conceived the study, developed the methodology and software, designed and conducted all experiments, analyzed and visualized the results, and wrote and revised the manuscript.

\subsubsection*{Acknowledgments}

This research was conducted using the Supermicro ARS-111GL-DNHR-LCC and FUJITSU Server PRIMERGY CX2550 M7 (Miyabi) at Joint Center for Advanced High Performance Computing (JCAHPC).
Additional computations were carried out using resources of the supercomputer Genkai offered under the category of General Projects by the Research Institute for Information Technology, Kyushu University.
This research also used computational resources of the supercomputer RIKYU provided by the RIKEN Center for Computational Science.

KMN is supported by the Center of Innovation for Sustainable Quantum AI (JST Grant Number JPMJPF2221).

This work was supported by RIKEN through its institutional funding.

\bibliographystyle{iclr2027_conference}
\bibliography{main}
\appendix


\clearpage
\section{Experimental Setup}\label[appendix]{apx:setup}

This section describes the primary Transformer and Multiscreen pretraining configurations under matched training data and token budgets.
Controlled comparisons and ablations document any deviations from this setup in their respective sections.

\paragraph{Tokenization.}
We adopt the GPT-2 tokenizer~\citep{radford2019language} with a vocabulary of 50,257 tokens.

\paragraph{Training Data.}
We pretrain the main-experiment models on the SlimPajama~\citep{cerebras2023slimpajama} dataset, a cleaned and deduplicated subset of RedPajama~\citep{weber2024redpajama}.
After tokenization, documents are packed back-to-back into fixed-length input sequences without inserting EOS tokens, while document boundaries are retained as metadata.
For every nonfinal token in a document, the prediction target is the next token in that document; for the final token, the target is EOS.
For example, documents $(a,b,c,d)$ and $(e,f,g,h,i)$ with sequence length $8$ produce input $(a,b,c,d,e,f,g,h)$ and targets $(b,c,d,\langle\mathrm{EOS}\rangle,f,g,h,i)$.
The corresponding document IDs are $(0,0,0,0,1,1,1,1)$ and are used only when document masking is enabled.
The tokenized documents contain approximately 627 billion tokens in total.
For these experiments, we use $2^{38}$ tokens (approximately 44\% of the dataset), obtained by partitioning the tokenized corpus into contiguous sequences of length $2^{12}$ and selecting $2^{26}$ sequences at random.
For validation, we use $2^{26}$ tokens from the SlimPajama validation split, packed into sequences of the same length $2^{12}$ as in training.

\paragraph{Document Masking.}
When document masking is enabled, each query can attend only to causal keys with the same document ID.
The primary Transformer and Multiscreen runs use this mask.
The controlled attention-mechanism comparison disables document masking for all variants because several fused kernels do not support it, as detailed in \cref{apx:attention-comparison}.

\paragraph{Packing Rationale.}
This construction assigns exactly one meaningful prediction target to every input token.
If EOS tokens were instead inserted into the input, each EOS position would either be trained to predict the first token of the next, unrelated document or be excluded from the cross-entropy loss.
The former trains the model to predict the first token of an unrelated document at an EOS position, although that token is not a meaningful continuation of the preceding document.
The latter makes the number of supervised targets per nominal token batch depend on the number of packed document boundaries.
In our construction, document-final tokens predict EOS and all other tokens predict the next token in the same document, so every input token is paired with exactly one meaningful supervised target.
With document masking, every prediction conditions only on tokens from the same document.
Without it, a token's causal context may include a preceding unrelated document, so the model is trained to predict within the current document in the presence of unrelated preceding context, while document-final tokens still mark document boundaries by predicting EOS.

\paragraph{Model Sizes.}
To analyze scaling behavior, we train both Transformer and Multiscreen models across multiple parameter scales.

For the Transformer baseline, we adopt a LLaMA-style architecture~\citep{touvron2023llama}, with architecture hyperparameters based on those used in Pythia~\citep{biderman2023pythia}, and apply weight tying between the input embedding and the language modeling head.
We follow Pythia configurations where applicable, including model sizes and peak learning rates.
Detailed configurations of the Transformer baseline are provided in \cref{apx:tf-config}.

For Multiscreen, the architecture hyperparameters and initialization scheme are provided in \cref{apx:musc-config,apx:init}.

\paragraph{Training Setup.}
All main-experiment models are pretrained with a sequence length of $2^{12}$ (4,096).
We use a global batch size of $2^{22}$ tokens.
Together with the $2^{38}$-token budget, this gives $2^{16}$ optimizer steps per run.

All models are optimized using AdamW~\citep{loshchilov2017decoupled} with $(\beta_1, \beta_2) = (0.9, 0.95)$ and $\epsilon = 10^{-8}$.
We linearly warm up the learning rate for $2^{12}$ steps, after which it is kept constant.
This warmup-then-constant schedule keeps the optimization protocol simple and comparable across architectures under a fixed token budget, and is consistent with recent work studying long stable or decay-free learning-rate phases~\citep{haegele2024scaling,wen2025understanding,yano2026pretraining}.
We train using bfloat16 automatic mixed precision.

Neither architecture uses weight decay or gradient clipping in the main experiments.
For Transformer, we selected this configuration because it achieved lower validation loss than the Pythia-style setting with weight decay $0.1$ and gradient clipping at $1.0$.
The learning-rate sweep in \cref{sec:lr} includes the latter setting as an explicit Transformer control.
For Transformer, we use RoPE with $\theta = 10{,}000$ and model-size-dependent learning rates based on Pythia~\citep{biderman2023pythia}.
The learning rate is set to the peak value used in Pythia for each model scale:
$1\times10^{-3}$ for 8M and 45M models, $3\times10^{-4}$ for 353M, and $2\times10^{-4}$ for 1.3B.

For Multiscreen, we use the same optimizer configuration as the Transformer baseline apart from the learning rate.
We use a learning rate of $2^{-4}$ (i.e., $0.0625$), substantially larger than those used for Transformer, enabled by the improved stability of Multiscreen demonstrated in \cref{sec:lr}.

\paragraph{Training Hardware.}
All scaling experiments use NVIDIA GH200 GPUs.
The 8M, 45M, and 353M Transformer models and the 4M, 28M, and 286M Multiscreen models use 16 GPUs, whereas both 1.3B models and the 4B Multiscreen model use 64 GPUs.

\paragraph{Implementations.}
Transformer uses \texttt{torch.nn.{\allowbreak}attention.{\allowbreak}varlen.{\allowbreak}varlen\_attn}, backed by PyTorch's built-in FlashAttention-2 kernel~\citep{dao2024flashattention2}, for both training and inference, while Multiscreen uses custom Triton kernels~\citep{tillet2019triton} tailored to its screening mechanism.
All experiments use PyTorch 2.13.0, CUDA 13.2, and Triton 3.7.1.
The hardware and protocol used for TTFT measurements are reported in \cref{apx:ttft}.

\clearpage
\section{Zero-Shot Downstream Evaluation}\label[appendix]{apx:downstream}

We evaluate the 1.3B Transformer and Multiscreen models in the zero-shot setting using version 0.4.12 of the Language Model Evaluation Harness~\citep{sutawika2026lmeval}, with \texttt{num\_fewshot=0} for every task.
We report normalized accuracy for HellaSwag~\citep{zellers2019hellaswag}, PIQA~\citep{bisk2020piqa}, ARC-Easy and ARC-Challenge~\citep{clark2018arc}, SciQ~\citep{welbl2017crowdsourcing}, and OpenBookQA~\citep{mihaylov2018openbookqa}.
We report ordinary accuracy for WinoGrande~\citep{sakaguchi2020winogrande}, BoolQ~\citep{clark2019boolq}, and MMLU~\citep{hendrycks2021measuring}.
The results are summarized in \cref{tb:downstream}, with the average computed as the unweighted mean across all nine tasks.

\begin{table}[h]
\centering
\caption{Zero-shot downstream accuracy (\%) of the 1.3B models, with the better result for each benchmark shown in bold.}
\label{tb:downstream}
\begin{tabular}{lrr}
\toprule
& \multicolumn{2}{c}{Accuracy (\%) $\uparrow$} \\
\cmidrule(lr){2-3}
Benchmark & Transformer & Multiscreen \\
\midrule
HellaSwag & 55.58 & \textbf{59.02} \\
PIQA & 72.91 & \textbf{73.99} \\
ARC-Easy & 52.82 & \textbf{55.60} \\
ARC-Challenge & 29.35 & \textbf{29.69} \\
WinoGrande & 58.41 & \textbf{59.43} \\
BoolQ & 61.80 & \textbf{64.77} \\
SciQ & \textbf{83.40} & 79.40 \\
OpenBookQA & 34.00 & \textbf{35.60} \\
MMLU & 23.42 & \textbf{25.53} \\
\midrule
Average & 52.41 & \textbf{53.67} \\
\bottomrule
\end{tabular}
\end{table}

\clearpage
\section{Multiscreen Architecture and Scaling}\label[appendix]{apx:musc-config}

We scale Multiscreen using a single supraparameter $\Psi$ that jointly determines the number of layers, the number of heads, and the embedding dimension.
In our experiments, the number of layers and the number of heads are set to $N_\text{L}=N_\text{H}=\Psi$, and the embedding dimension is set to $d_\text{E}=\Psi^2$.

Other architectural hyperparameters are kept fixed across model scales, as summarized in \cref{tb:musc-hp-base}.
The resulting model configurations and parameter counts across scales are shown in \cref{tb:musc-hp}.

\begin{table}[ht]
\caption{Architectural hyperparameters of Multiscreen used in our experiments.}
\label{tb:musc-hp-base}
\centering
\begin{tabular}{lcc}
\toprule
\textbf{Hyperparameter} & \textbf{Symbol} & \textbf{Value / scaling} \\
\midrule
Number of layers & $N_\text{L}$ & $\Psi$ \\
Number of heads & $N_\text{H}$ & $\Psi$ \\
Embedding dimension & $d_\text{E}$ & $\;\Psi^2$ \\
Query/key dimension & $d_\text{K}$ & 16 \\
Value dimension & $d_\text{V}$ & 64 \\
MiPE threshold & $w_\text{th}$ & 256 \\
\bottomrule
\end{tabular}
\end{table}

\begin{table}[ht]
\caption{Architecture configurations and parameter counts for the Multiscreen models used in the scaling experiments.}
\label{tb:musc-hp}
\centering
\begin{tabular}{@{}l@{\hspace{9pt}}c@{\hspace{9pt}}c@{\hspace{9pt}}c@{\hspace{9pt}}c@{\hspace{9pt}}c@{}}
\toprule
\textbf{Quantity} & \textbf{4M} & \textbf{28M} & \textbf{286M} & \textbf{1.3B} & \textbf{4B} \\
\midrule
Supraparameter ($\Psi$) & 8 & 16 & 32 & 48 & 64 \\
Total params & 4,134,146 & 27,546,626 & 286,347,266 & 1,304,884,226 & 3,963,961,346 \\
Non-embedding params & 917,698 & 14,680,834 & 234,884,098 & 1,189,092,098 & 3,758,108,674 \\
\bottomrule
\end{tabular}
\end{table}

\clearpage
\section{Transformer Baseline Configurations}\label[appendix]{apx:tf-config}

We describe the architecture configurations of the Transformer baseline models used in our experiments.
The Transformer baseline adopts a LLaMA-style architecture~\citep{touvron2023llama}, with architecture hyperparameters largely aligned with those used in Pythia~\citep{biderman2023pythia}.
We apply weight tying between the input embedding and the language-modeling head, which differs from the original Pythia configuration, to improve parameter efficiency and strengthen the baseline.

The head dimension is defined as $d_\text{E}/N_\text{H}$, and the feed-forward dimension is set to $\lfloor \frac{8}{3}d_\text{E} \rfloor$.
The architecture configurations across model scales are summarized in \cref{tb:tf-hp}.

\paragraph{Initialization.}
For the Transformer baseline, we follow the Pythia/GPT-NeoX initialization scheme~\citep{biderman2023pythia,black2022gptneox}.
Specifically, all weight matrices except residual output projections are initialized using small initialization with standard deviation $\sqrt{2/(5d_\text{E})}$, while residual output projections (i.e., the output projections of the attention and feed-forward blocks) are initialized using Wang initialization with standard deviation $2/(N_\text{L}\sqrt{d_\text{E}})$.

\begin{table}[ht]
\caption{Architecture configurations and parameter counts for the Transformer baseline models used in the scaling experiments.}
\label{tb:tf-hp}
\centering
\begin{tabular}{lcccc}
\toprule
\textbf{Quantity} & \textbf{8M} & \textbf{45M} & \textbf{353M} & \textbf{1.3B} \\
\midrule
Number of layers ($N_\text{L}$) & 6 & 6 & 24 & 24 \\
Number of heads ($N_\text{H}$) & 4 & 8 & 16 & 16 \\
Embedding dimension ($d_\text{E}$) & 128 & 512 & 1,024 & 2,048 \\
Total params & 7,613,312 & 44,609,024 & 353,453,056 & 1,310,935,040 \\
Non-embedding params & 1,180,416 & 18,877,440 & 301,989,888 & 1,208,008,704 \\
\bottomrule
\end{tabular}
\end{table}

\clearpage
\section{Multiscreen Initialization}\label[appendix]{apx:init}

Except for the gate projection $W_\text{G}$, we initialize all projection matrices and the embedding matrix from zero-mean Gaussian distributions whose standard deviations are scaled inversely with the square root of their output dimension (i.e., the second dimension of each weight matrix), as summarized in \cref{tb:init_hp}.
The gate projection $W_\text{G}$ is initialized with a fixed scale independent of dimension.
For matrices whose outputs are subsequently unit-normalized, the global initialization scale does not affect the norm of the normalized vectors, but can still affect early optimization dynamics; for projections not followed by such normalization, it directly affects the initial activation scale.

The window parameter $s_\text{w}$ is initialized independently in each layer with values linearly spaced across heads from $0$ to $\log w_\text{th}$.
The parameter $s_\text{r}$ is initialized to zero, corresponding to an initial acceptance width of $r = 0.5$.
The residual output scale $s_\text{O}$ is initialized to approximately normalize the total contribution across all tiles in the model, ensuring that the aggregated residual updates remain well-scaled as the number of tiles increases.

The parameters $s_\text{E}$ and $s_\text{F}$ control the scale of input embeddings and output logits, respectively, and are initialized to set appropriate scales for these quantities.

\begin{table}[ht]
\caption{
Parameter shapes and initialization of Multiscreen.
All matrix initializations are zero-mean Gaussian, and $\sigma$ denotes their standard deviation.
}
\label{tb:init_hp}
\centering
\begin{tabular}{ccc}
\toprule
\textbf{Parameter} & \textbf{Shape} & \textbf{Initialization} \\
\midrule
$W_\text{Q}$ & $(d_\text{E}, d_\text{K})$ & $\sigma=0.1/\sqrt{d_\text{K}}$ \\
$W_\text{K}$ & $(d_\text{E}, d_\text{K})$ & $\sigma=0.1/\sqrt{d_\text{K}}$ \\
$W_\text{V}$ & $(d_\text{E}, d_\text{V})$ & $\sigma=0.1/\sqrt{d_\text{V}}$ \\
$W_\text{G}$ & $(d_\text{E}, d_\text{V})$ & $\sigma=0.1$ \\
$W_\text{O}$ & $(d_\text{V}, d_\text{E})$ & $\sigma=0.1/\sqrt{d_\text{E}}$ \\
$W_\text{E}$ & $(|\mathcal{V}|, d_\text{E})$ & $\sigma=0.1/\sqrt{d_\text{E}}$ \\
$s_\text{w}$ & per-tile scalar & evenly spaced over $[0,\log w_\text{th}]$ per layer \\
$s_\text{r}$ & per-tile scalar & $0$ \\
$s_\text{O}$ & per-tile scalar & $\log\bigl(1 / \sqrt{N_\text{H}N_\text{L}}\bigr)$ \\
$s_\text{E}$ & global scalar & $0$ \\
$s_\text{F}$ & global scalar & $\log \sqrt{d_\text{E}}$ \\
\bottomrule
\end{tabular}
\end{table}

\clearpage
\section{Additional Scaling Analysis}\label[appendix]{apx:scaling}

\begin{figure}[ht]
\centering
\includegraphics[width=0.55\columnwidth]{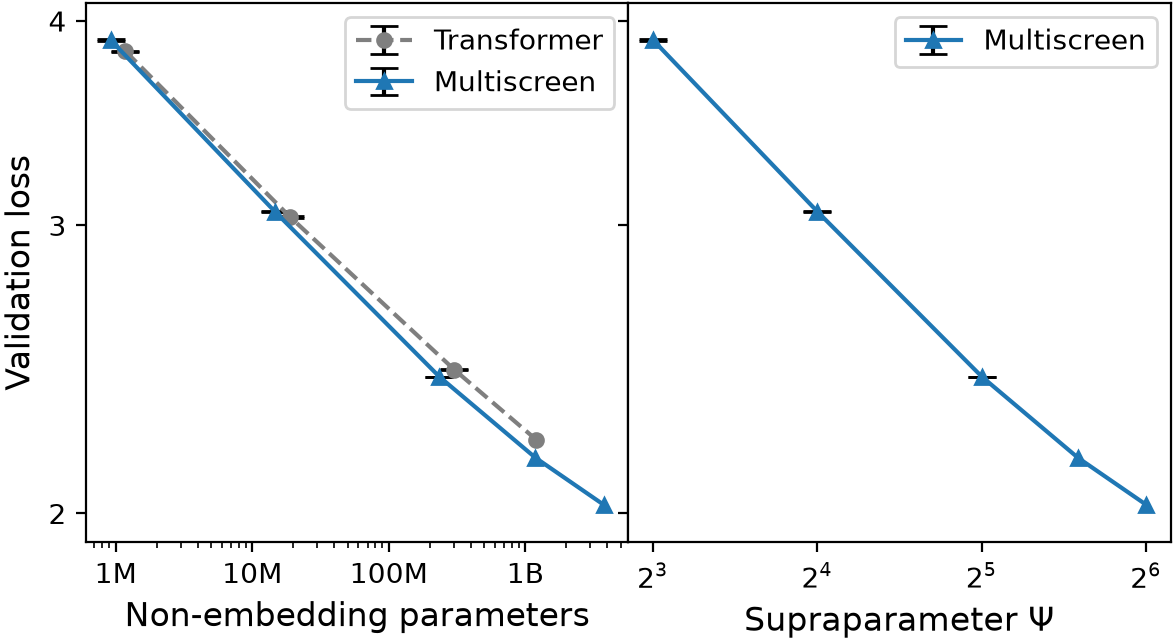}
\caption{
Scaling behavior under alternative definitions of model size.
Left: scaling behavior of Transformer and Multiscreen with respect to non-embedding parameters.
Right: scaling behavior of Multiscreen with respect to the supraparameter $\Psi$.
Error bars indicate one standard deviation across independently trained models where multiple runs are available.
}
\label{fig:scaling-alt}
\end{figure}

We further analyze scaling behavior using alternative definitions of model size, as shown in \cref{fig:scaling-alt}.

\paragraph{Non-embedding parameters.}
In the left panel of \cref{fig:scaling-alt}, we plot validation loss as a function of non-embedding parameters for both Transformer and Multiscreen.
The points follow a visually more linear trend on the log--log axes under this parameterization, while the relative advantage of Multiscreen over Transformer is preserved.

\paragraph{Supraparameter $\Psi$.}
In the right panel of \cref{fig:scaling-alt}, we analyze scaling behavior using a unified supraparameter $\Psi$ defined for Multiscreen.
Plotting against $\Psi$ yields a similarly regular empirical trend, suggesting that $\Psi$ provides a compact and unified characterization of model scaling for Multiscreen.

\begin{figure}[ht]
\centering
\includegraphics[width=0.65\linewidth]{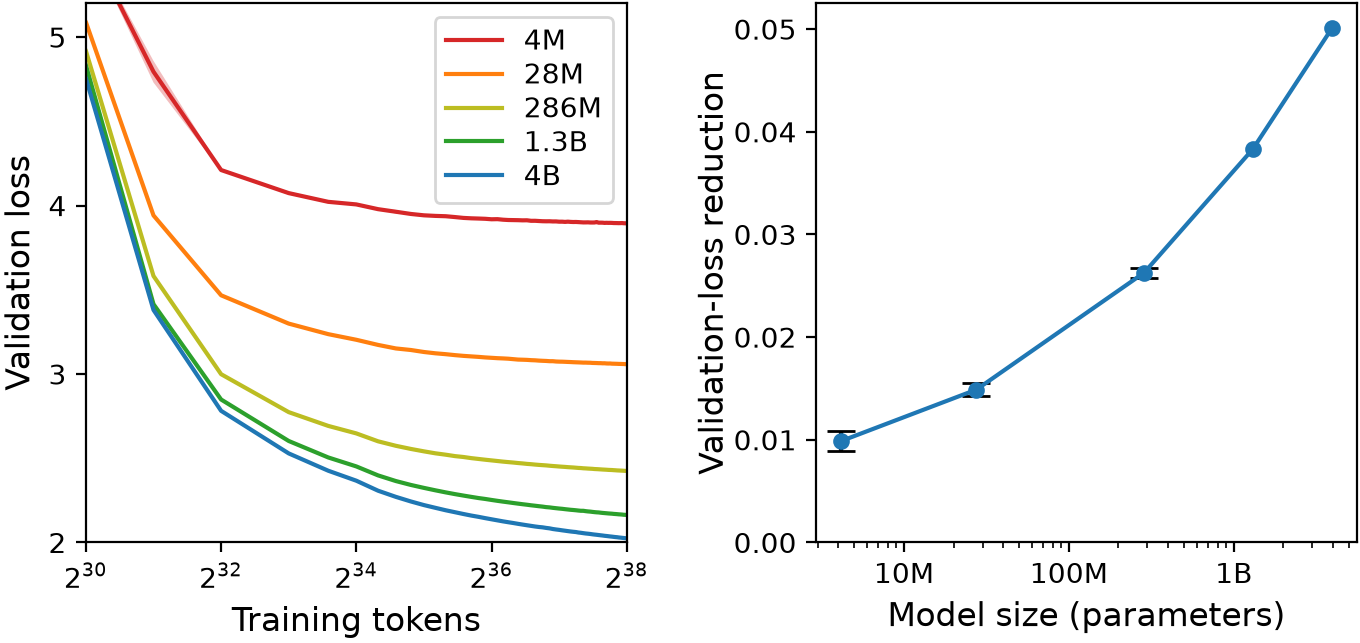}
\caption{
Validation-loss trajectories of Multiscreen models trained under the fixed $2^{38}$-token budget.
Left: validation loss as a function of training tokens.
Right: validation-loss reduction over the second half of training.
For the 4M, 28M, and 286M models, curves and shaded regions in the left panel show the mean and one standard deviation across three independently trained models, and error bars in the right panel indicate one standard deviation across the same runs.
The 1.3B and 4B results each use a single trained model.
}
\label{fig:valloss-curve}
\end{figure}

The empirical trend remains regular through 1.3B, while the 4B point shows a modest deviation in both parameterizations.
Because every model is trained on the same fixed token budget, the number of training tokens per parameter decreases with model size.
To assess whether incomplete convergence contributes to the 4B deviation, we examine validation loss throughout training in \cref{fig:valloss-curve}.
All model sizes continue to improve through $2^{38}$ training tokens, and the validation-loss reduction over the second half of training increases monotonically with model size.
Together, these observations suggest that the 4B deviation partly reflects incomplete convergence under the fixed token budget.

\clearpage
\section{Training Loss Dynamics}\label[appendix]{apx:lr-log}

To complement the learning-rate sweep in \cref{fig:lr-sweep}, we visualize training-loss trajectories from the same runs at representative learning rates.

\begin{figure}[ht]
\centering
\vspace{6mm}
\includegraphics[width=\linewidth]{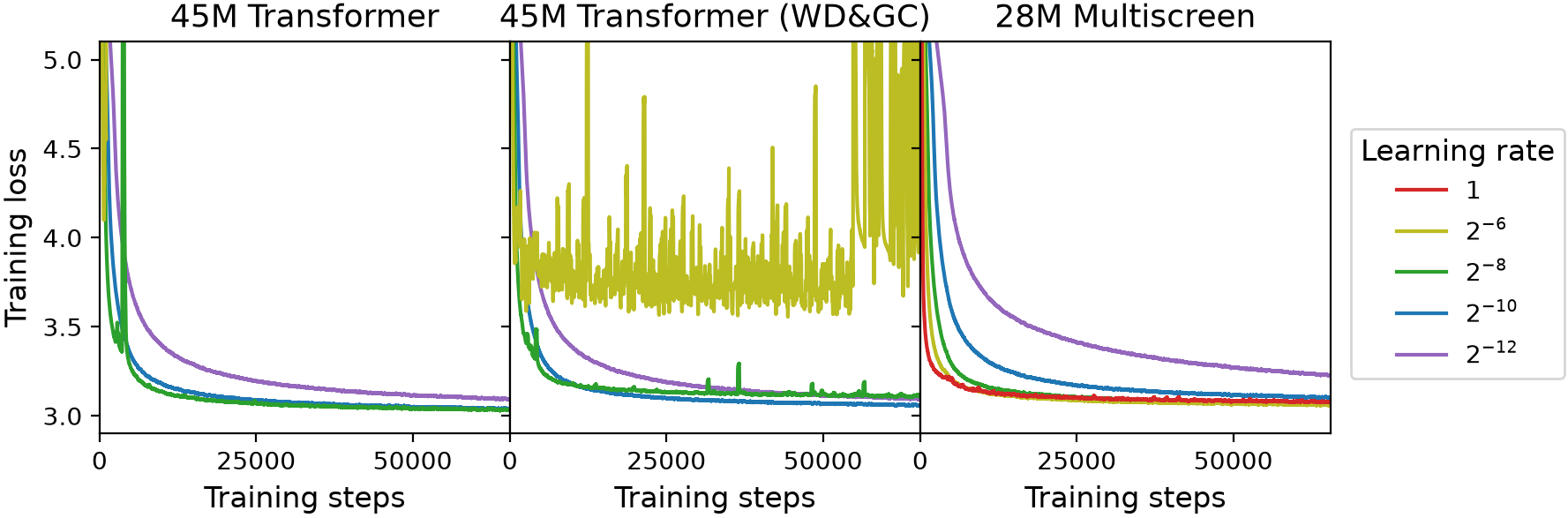}
\caption{
Training loss trajectories from the same runs as in \cref{fig:lr-sweep}, shown for representative learning rates.
Left: 45M Transformer without weight decay or gradient clipping; center: the Transformer control with weight decay $0.1$ and gradient clipping at $1.0$; right: 28M Multiscreen without either.
Curves are smoothed using a moving average over 256 training steps.
Divergent Transformer runs may lie entirely outside the displayed loss range.
}
\label{fig:lr-log}
\end{figure}

As shown in \cref{fig:lr-log}, at higher learning rates, both Transformer configurations either exhibit pronounced loss fluctuations or lie outside the plotted loss range, whereas Multiscreen follows smooth training trajectories and completes training even at learning rate $1$.

While \cref{fig:lr-sweep} summarizes final validation loss and divergence outcomes, the training trajectories shown here illustrate how instability emerges during optimization.
To better understand these behaviors, we next analyze the gradient dynamics in \cref{apx:grad-norm}.

\clearpage
\section{Gradient Norm Dynamics}\label[appendix]{apx:grad-norm}

We compare global gradient norms from one 1.3B Transformer run and one 1.3B Multiscreen run used in the scaling experiments (\cref{sec:scaling}).
The Transformer and Multiscreen models use learning rates $2\times10^{-4}$ and $2^{-4}$, respectively, with Multiscreen remaining stable at the substantially larger value.

\begin{figure}[ht]
\centering
\includegraphics[width=0.35\columnwidth]{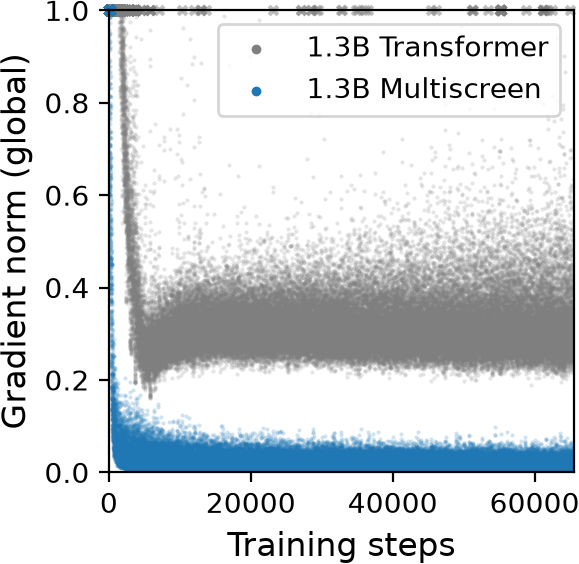}
\caption{
Global gradient-norm dynamics during training for the 1.3B Transformer and Multiscreen models used in the scaling experiments.
For visualization, values above 1 are clipped and shown with $\times$ markers.
}
\label{fig:grad-norm}
\end{figure}

As shown in \cref{fig:grad-norm}, Transformer maintains a nonzero lower floor relative to the spread of its gradient norms, whereas Multiscreen has a dense concentration near zero relative to its full gradient-norm range.
This contrast motivates testing whether a similar near-zero lower tail appears when controlled Transformer variants are stabilized at learning rate $1$.

\clearpage
\section{Components of Large-Learning-Rate Stability}\label[appendix]{apx:lr-components}

We conduct controlled diagnostic experiments at learning rate $1$ to determine whether the stability of Multiscreen is caused by screening itself or by its normalization and scaling components.
Every diagnostic run is trained for $2^{36}$ tokens ($2^{14}$ steps).
For the Transformer variants, ``QKNorm'' denotes an RMSNorm-based variant of query--key normalization~\citep{henry2020query} with a learned elementwise scale applied to queries and keys.
Here, ``PLNorm'' denotes pre-logit normalization, while ``PreNorm'' denotes the RMSNorm applied at the beginning of each Transformer layer.
Embedding normalization and PLNorm introduce no learned elementwise scales and only project their inputs onto the unit hypersphere.
Together with the learned input and logit scales $\mathrm{e}^{s_\text{E}}$ and $\mathrm{e}^{s_\text{F}}$, they form the SIO parameterization defined in \cref{sec:model}.
The learned scalars $s_\text{E}$, $s_\text{F}$, and $s_\text{O}$ parameterize the input-embedding, output-logit, and residual-output scales, respectively.

\begin{table}[ht]
\centering
\caption{
Training outcomes at learning rate $1$ under controlled modifications to Transformer and Multiscreen.
``Stable'' indicates completion of the $2^{14}$-step diagnostic run without divergence.
``SIO $-$ PLNorm'' denotes the SIO components other than PLNorm.
}
\label{tab:lr-components}
\begin{tabular}{lc}
\toprule
Configuration & Outcome \\
\midrule
Transformer & Diverged \\
Transformer $+$ SIO & Diverged \\
Transformer $+$ PLNorm & Diverged \\
Transformer $+$ SIO $-$ PLNorm & Diverged \\
Transformer $+$ QKNorm & Diverged \\
Transformer $+$ QKNorm $+$ SIO & Stable \\
Transformer $+$ QKNorm $+$ PLNorm & Stable \\
Transformer $+$ QKNorm $+$ SIO $-$ PLNorm & Stable \\
Transformer $-$ PreNorm $+$ QKNorm $+$ SIO & Diverged \\
Transformer $-$ PreNorm $+$ QKNorm $+$ PLNorm & Diverged \\
Transformer $-$ PreNorm $+$ QKNorm $+$ SIO $-$ PLNorm & Diverged \\
\midrule
Multiscreen & Stable \\
Multiscreen $-$ SIO & Diverged \\
Multiscreen $-$ PLNorm & Stable \\
Multiscreen $-$ (SIO $-$ PLNorm) & Diverged \\
Multiscreen $-$ $s_\text{O}$ & Diverged \\
Multiscreen $+$ PreNorm & Diverged \\
\bottomrule
\end{tabular}
\end{table}

The clearest shared pattern in \cref{tab:lr-components} is the interaction between Q/K normalization and SIO.
For Transformer, QKNorm and SIO each diverge when introduced alone, whereas their combination is stable at learning rate $1$.
Multiscreen likewise incorporates both Q/K unit-length normalization and SIO by design; the standard model is stable at learning rate $1$, whereas removing SIO causes divergence.
Together, these results suggest that pairing a form of Q/K normalization with SIO contributes strongly to large-learning-rate stability across both architectures.
Q/K unit-length normalization is not independently ablated in Multiscreen because it establishes the fixed $[-1,1]$ input scale assumed by Trim and is therefore inseparable from the screening-unit design.

The remaining controls refine this shared pattern.
For Transformer, QKNorm also stabilizes both PLNorm and SIO without PLNorm when PreNorm is retained.
Notably, PreNorm plays opposite roles across the two architectures: removing it from any of the three stable Transformer configurations causes divergence, whereas adding it to Multiscreen also causes divergence.
For Multiscreen, removing PLNorm alone preserves stability, whereas retaining only PLNorm from SIO causes divergence.
Removing $s_\text{O}$ also causes Multiscreen to diverge.
Thus, the precise supporting components and normalization placement remain architecture dependent.

\begin{figure}[t]
\centering
\includegraphics[width=0.75\linewidth]{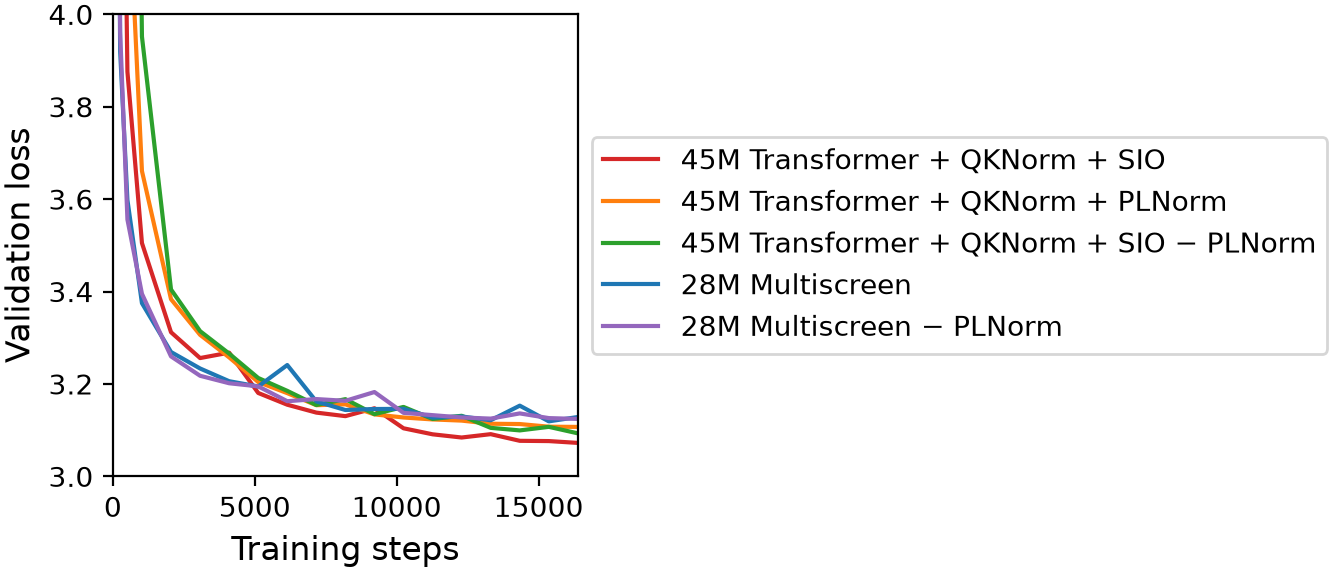}
\caption{
Validation-loss trajectories for five selected configurations that remain stable at learning rate $1$.
PLNorm denotes pre-logit normalization, and each curve shows one $2^{14}$-step training run.
}
\label{fig:lr1-valloss}
\end{figure}

\begin{figure}[t]
\centering
\includegraphics[width=\linewidth]{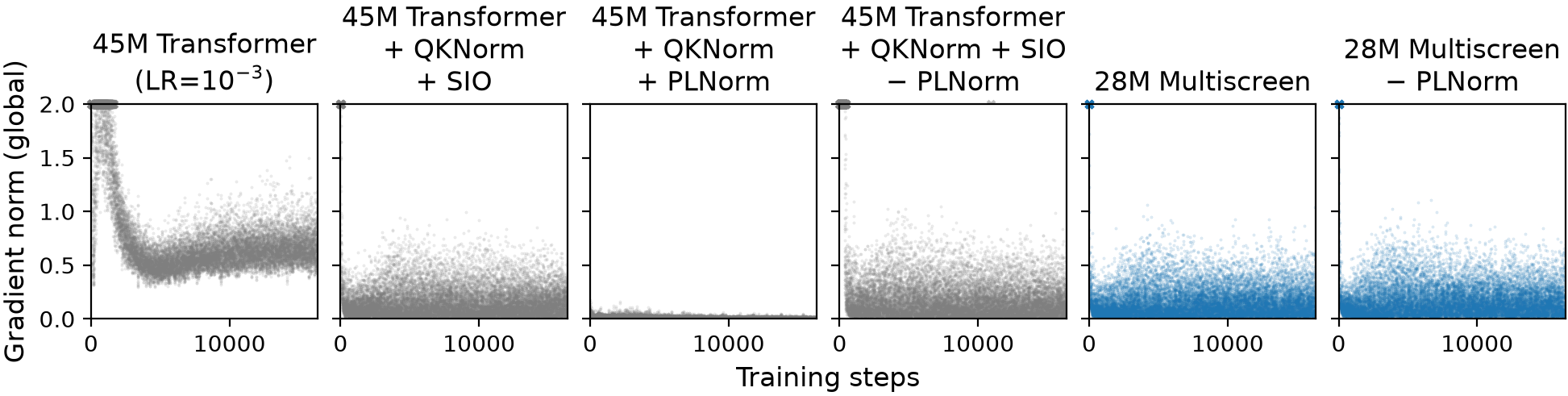}
\caption{
Global gradient norms for five selected configurations that remain stable at learning rate $1$, together with the standard 45M Transformer at learning rate $10^{-3}$ as a low-learning-rate reference.
PLNorm denotes pre-logit normalization, and each panel shows one $2^{14}$-step training run.
For visualization, values above 2 are clipped and shown with $\times$ markers.
}
\label{fig:lr1-grad}
\end{figure}

The five selected stable configurations exhibit broadly similar validation-loss trajectories in this diagnostic run (\cref{fig:lr1-valloss}).
Although their absolute scales and dispersions differ substantially, all five exhibit a dense lower tail near zero relative to their overall gradient-norm ranges, whereas the standard Transformer at learning rate $10^{-3}$ maintains a visibly nonzero lower floor (\cref{fig:lr1-grad}).
This controlled comparison reproduces the contrast in \cref{fig:grad-norm} and reveals a common gradient-norm signature of large-learning-rate stability across architectures.

Taken together, these experiments identify Q/K normalization paired with SIO as a transferable design pattern for large-learning-rate stability, while showing that its supporting component interactions depend on the backbone.

\clearpage
\section{Visualization of Distance-Aware Relevance}\label[appendix]{apx:relevance-map}

\begin{figure}[hp]
\centering
\includegraphics[width=\linewidth]{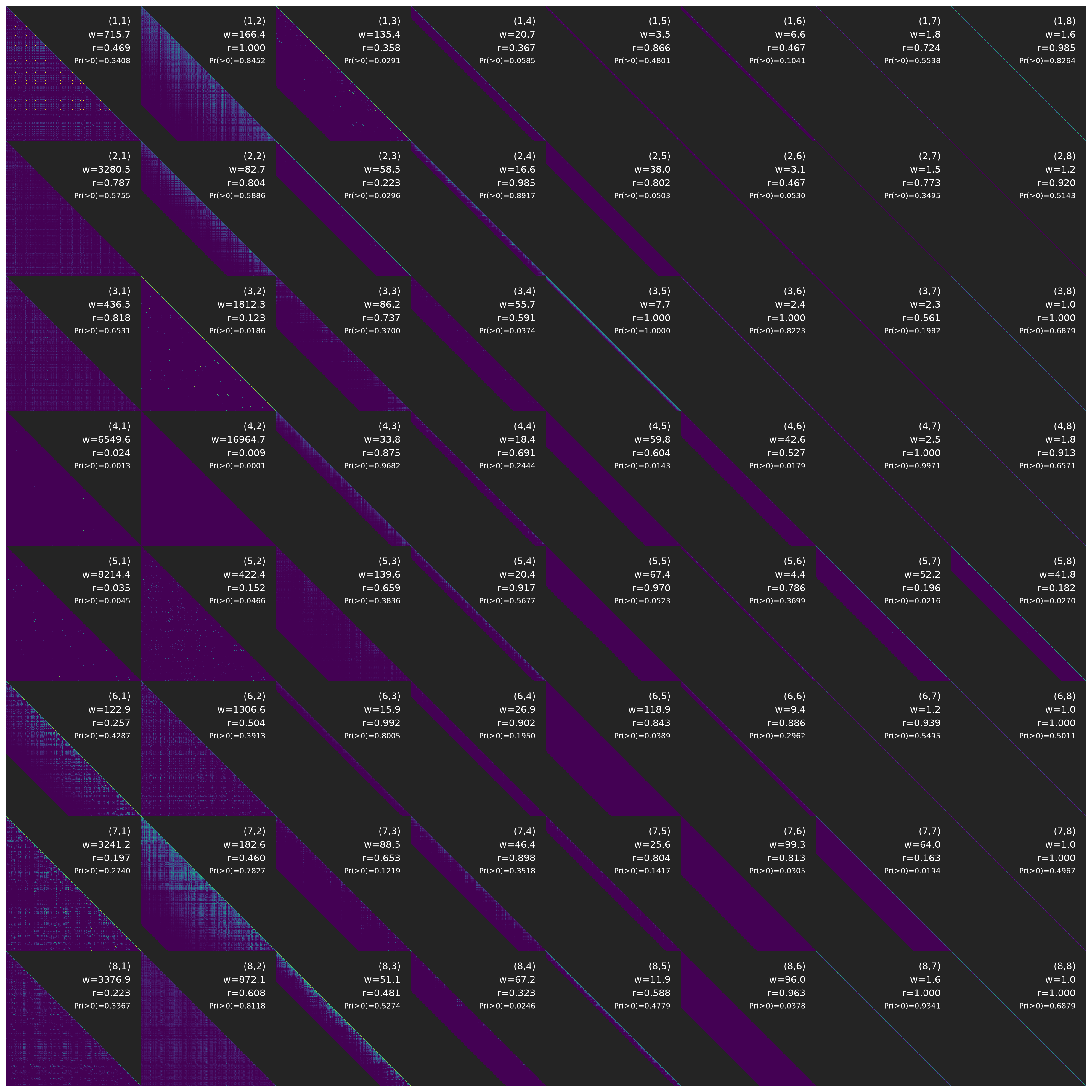}\\
\hfill
\includegraphics[width=0.3\linewidth]{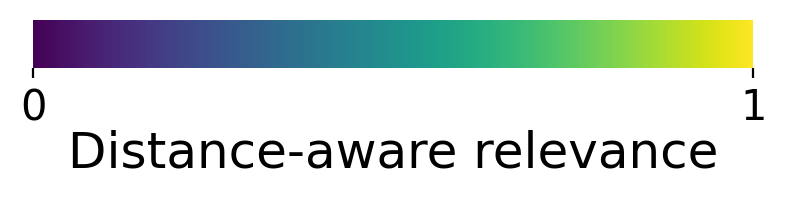}
\caption{
Distance-aware relevance maps across layers and heads.
Each map shows the distance-aware relevance $\alpha^\text{d}_{ij}$, with rows indexing query positions and columns indexing key positions.
Dark gray regions indicate causally inaccessible positions or positions outside the learned screening window.
Each tile is annotated with its layer and head indices, the learned screening window $w$, the acceptance width $r$, and the fraction of nonzero relevance values $\Pr(\alpha^\text{d}_{ij} > 0)$ among causally accessible positions within the window, providing a summary of its sparsity and coverage.
}
\label{fig:relevance-map}
\end{figure}

We visualize distance-aware relevance maps $\alpha^\text{d}_{ij}$ for all tiles in a 4M Multiscreen model used in \cref{sec:scaling}, as shown in \cref{fig:relevance-map}.
These visualizations highlight the presence of learned screening windows and the extent of sparsity within them.

The input sequence is a 228-token natural-language passage (the abstract of this paper).
All layers and heads are shown, arranged in a grid layout where rows correspond to layers and columns correspond to heads.

Within each map, dark gray regions indicate causally inaccessible positions or positions outside the learned screening window, while color intensity represents the magnitude of $\alpha^\text{d}_{ij}$ over the remaining positions.
Each tile is annotated with its layer and head indices, the learned screening window $w$, the acceptance width $r$, and the fraction of nonzero relevance values among causally accessible positions within the window, $\Pr(\alpha^\text{d}_{ij} > 0)$.

Different tiles cover different context ranges, with some focusing on local neighborhoods and others covering broader portions of the sequence.
Many maps contain a substantial number of zero entries, with the degree of sparsity varying across tiles.

\clearpage
\section{Distribution of Learned Screening Parameters}\label[appendix]{apx:wr}

Each screening unit has independently learned scalar parameters $s_\text{w}$ and $s_\text{r}$, which determine the screening window $w$ and acceptance width $r$.
To better understand how these parameters are used after training, we visualize the resulting values of $w$ and $r$ across all layers and heads of the 4B Multiscreen model.
The figure reports the raw finite values of $w$ before applying the default inference rule, which replaces values exceeding the pretraining context length with $w=\infty$.

\begin{figure}[h]
\centering
\includegraphics[width=0.5\linewidth]{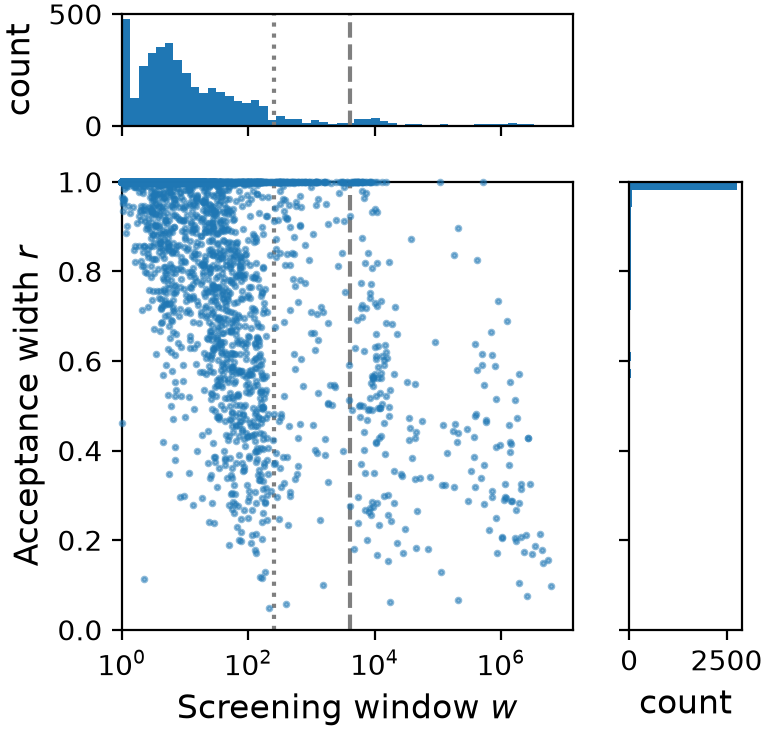}
\caption{
Distribution of learned screening parameters in the 4B Multiscreen model.
Each point corresponds to one screening unit, with horizontal position indicating the learned screening window $w$ and vertical position indicating the learned acceptance width $r$.
The horizontal axis is shown on a logarithmic scale.
Marginal histograms show the corresponding distributions of $w$ and $r$.
The dotted vertical line marks the MiPE threshold ($w_\text{th}=2^8$), at or above which MiPE becomes inactive, and the dashed vertical line marks the pretraining context length ($2^{12}$), beyond which the default inference rule sets $w=\infty$.
Raw learned windows span several orders of magnitude, with many local or medium-range units and a smaller number exceeding the pretraining context length.
}
\label{fig:wr}
\end{figure}

As shown in \cref{fig:wr}, learned screening windows span several orders of magnitude.
In this model, most units remain local or medium-range, while $6.0\%$ learn window values exceeding the pretraining context length and are therefore treated as having $w=\infty$ under the default inference rule.
The learned acceptance widths are often close to one, but also vary substantially across units, indicating that different screening units specialize to different degrees of selectivity.
This diversity suggests that Multiscreen combines many local screening units with a smaller number of broad-context units.

\clearpage
\section{Long-Context Perplexity: Evaluation Details}\label[appendix]{apx:lcp}

We provide additional details for the long-context perplexity evaluation in \cref{sec:lcp}, along with robustness results using smaller models trained with multiple random seeds.

\paragraph{Evaluation setup.}
We evaluate long-context language modeling using position-dependent perplexity over long sequences.
At each reported position, perplexity is averaged over a trailing window spanning the preceding $10\%$ of the context length and ending at that position to reduce local variance.

We use the PG-19 dataset~\citep{rae2020compressive}, consisting of books published before 1919 from Project Gutenberg.\footnote{\url{https://huggingface.co/datasets/Geralt-Targaryen/pg19}}
We pool the training, validation, and test splits and use all 5,747 documents whose tokenized length exceeds $2^{17}$.
The test split alone contains too few documents of this length to provide stable position-dependent estimates at the longest contexts.
PG-19 was not explicitly deduplicated against SlimPajama, so overlap with the pretraining corpus may remain.
We therefore use this evaluation as a controlled comparison between architectures rather than as a held-out estimate of absolute generalization performance.
Every compared model encountered exactly the same pretraining tokens in the same order and is evaluated on the same segments, preventing differential data exposure between Transformer and Multiscreen.
For each document, we extract a contiguous segment of $2^{17}+1$ tokens centered at the midpoint of the document.
This construction ensures that predictions are evaluated in the middle of long contexts rather than near document boundaries.

\paragraph{YaRN scaling.}
For Transformer, we apply YaRN~\citep{peng2024yarn} with multiple scaling factors to assess extrapolation beyond the training context length.
For base models trained with sequence length $2^{12}$, we evaluate scaling factors
$\times1$, $\times2$, $\times4$, $\times8$, $\times16$, $\times32$, and $\times64$.
Multiscreen has no RoPE component, so YaRN scaling does not apply.

\paragraph{Seed-averaged smaller-model results.}
In addition to the 1.3B results in \cref{fig:lcp}, we evaluate the 353M Transformer and 286M Multiscreen models used in the scaling experiments.
For each architecture, we use three independently trained models with different random seeds and report the mean and one standard deviation.

\begin{figure}[ht]
\centering
\includegraphics[width=0.6\linewidth]{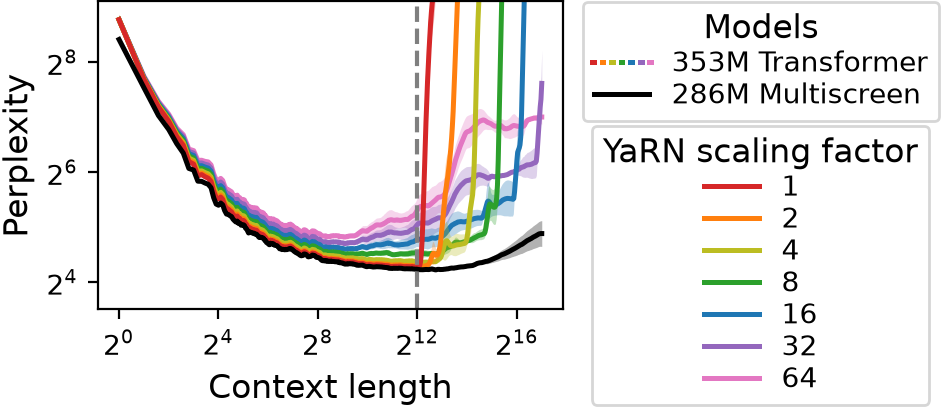}
\caption{
Long-context perplexity comparison for 353M Transformer and 286M Multiscreen models.
Curves show mean perplexity over three independently trained models, and shaded regions indicate one standard deviation.
The horizontal axis indicates the context length at which position-dependent perplexity is evaluated, and the vertical axis is perplexity.
The dashed vertical line marks the pretraining context length ($2^{12}$).
Colored curves show Transformer with different YaRN scaling factors, while the black curve shows Multiscreen, which has no RoPE component and therefore requires no YaRN scaling.
Multiscreen maintains stable perplexity substantially beyond the training context, while Transformer exhibits a tradeoff between delayed breakdown and increased perplexity at shorter contexts.
}
\label{fig:lcp-small}
\end{figure}

As shown in \cref{fig:lcp-small}, the same qualitative behavior as in \cref{sec:lcp} is observed.
Multiscreen maintains stable perplexity substantially beyond the training context, whereas Transformer with smaller YaRN scaling factors exhibits sharp degradation beyond the corresponding scaled range.
Increasing the YaRN scaling factor delays this degradation but increases perplexity at shorter contexts, including within the training range.

\clearpage
\section{ABCDigits: Task Construction and Evaluation Details}\label[appendix]{apx:abcd}

We provide additional construction and evaluation details for ABCDigits, the synthetic key--value retrieval benchmark evaluated in \cref{fig:abcd} and discussed in \cref{sec:abcd}.

\paragraph{Motivation.}
ABCDigits builds on synthetic associative recall and key--value retrieval benchmarks studied in prior work~\citep{graves2014neural,olsson2022context,liu2024lost,arora2024zoology}.
We cast retrieval as a structured completion problem that directly requires recovering the associated value.
This provides a direct probe of retrieval behavior while reducing confounding effects from natural-language semantics, instruction following, and prompt-specific factors.

\paragraph{Task construction.}
Each ABCDigits instance presents a shuffled list of equations mapping uppercase letters to six-digit integers, such as \texttt{A=967892}.
A query is formed by appending a target letter followed by an equals sign, such as \texttt{L=}, and the model must complete the corresponding integer.
The target mapping appears exactly once in the context, requiring the model to locate this unique occurrence.
Since each letter is consistently mapped to a single integer across the context, the task implicitly enforces a one-to-one key--value correspondence without requiring explicit instructions.

A key design choice is that the number of distinct keys remains fixed at 26 uppercase letters regardless of context length.
This removes the bias that retrieval becomes harder simply because the number of keys increases, allowing us to better isolate length-dependent retrieval behavior.

To ensure that extremely low-frequency equations are not anomalous within the context, we construct the non-target portion of the context in two stages.
First, we include exactly one instance of each of the 25 non-target equations.
We then fill the remaining context by sampling additional non-target equations from a highly skewed categorical distribution with weights proportional to $2^0,2^1,\ldots,2^{24}$, where the assignment from letters to weights is randomized for each instance.
This avoids making infrequent key--value pairs appear anomalous, ensuring that a range of frequencies is naturally represented within the context.
All non-target equations are then shuffled.
Finally, the unique target equation, such as \texttt{L=169428}, is inserted at a specified depth, and the query suffix \texttt{L=} is appended.
A concrete example of the resulting prompt is shown in \cref{fig:prompt} in the main text.

\paragraph{Evaluation protocol.}
Following the visualization protocol commonly used in needle-in-a-haystack evaluations~\citep{kamradt2023niah,arize2023niah}, we measure retrieval accuracy across a grid of context lengths and target depths.
Target context lengths range from $2^{12}$ to $2^{17}$, and depths are set to $0.1$, $0.3$, $0.5$, $0.7$, and $0.9$.
Because each equation is retained in full, the realized token count may differ slightly from its target.
Depth is implemented at the equation level: a depth of $d$ means that the target equation is inserted after approximately a fraction $d$ of the equations in the context, before the final query.

For each context length and depth, we generate 1,000 independent ABCDigits instances.
Every model is evaluated on the same instances at each setting.
We evaluate exact-match accuracy under greedy decoding with temperature $0$.
An instance is counted as correct only if the generated integer exactly matches the target value.

\paragraph{Models and inference settings.}
The main ABCDigits evaluation compares Transformer models with 8M, 45M, 353M, and 1.3B parameters against Multiscreen models with 4M, 28M, 286M, and 1.3B parameters.
For all models below 1.3B parameters, results are averaged over three independently trained models, with 1,000 instances evaluated per model and setting.
For the 1.3B models, we evaluate a single trained model for each architecture due to computational constraints.

For Transformer, we apply YaRN~\citep{peng2024yarn} with a scaling factor matched to each evaluation context length.
Specifically, for evaluation context length $L$ and training context length $L_{\text{train}}=2^{12}$, we set the YaRN scaling factor to $L/L_{\text{train}}$; for example, evaluation at $L=2^{15}$ uses a scaling factor of $8$.
Multiscreen has no RoPE component, so YaRN scaling does not apply.
Multiscreen's default inference rule is fixed across context lengths: after training, any learned screening window exceeding the maximum sequence length seen during training is converted once to $w=\infty$, and all windows then remain unchanged.

\clearpage
\section{Passkey Retrieval Results}\label[appendix]{apx:passkey}

We evaluate retrieval performance on the passkey retrieval task introduced by \citet{mohtashami2023random}, as a complementary test to ABCDigits.

\paragraph{Setup.}
The experimental setup follows that of ABCDigits (\cref{apx:abcd}).
We use the same range of context lengths ($2^{12}$ to $2^{17}$), target depths ($0.1$, $0.3$, $0.5$, $0.7$, and $0.9$), and number of trials (1,000 instances per setting).
As in the original task, each passkey is an integer sampled from 1 to 50,000; unlike its random-insertion protocol, we place the passkey at each fixed target depth in our evaluation grid.
We count an output as correct only when it produces the passkey immediately at the answer position, rather than searching the first 100 generated tokens for the first integer.
We use the same model scales and architecture-specific inference settings; only the prompt format is changed to follow the passkey retrieval task.

\begin{figure}[ht]
\centering
\vspace{6mm}
\includegraphics[width=\linewidth]{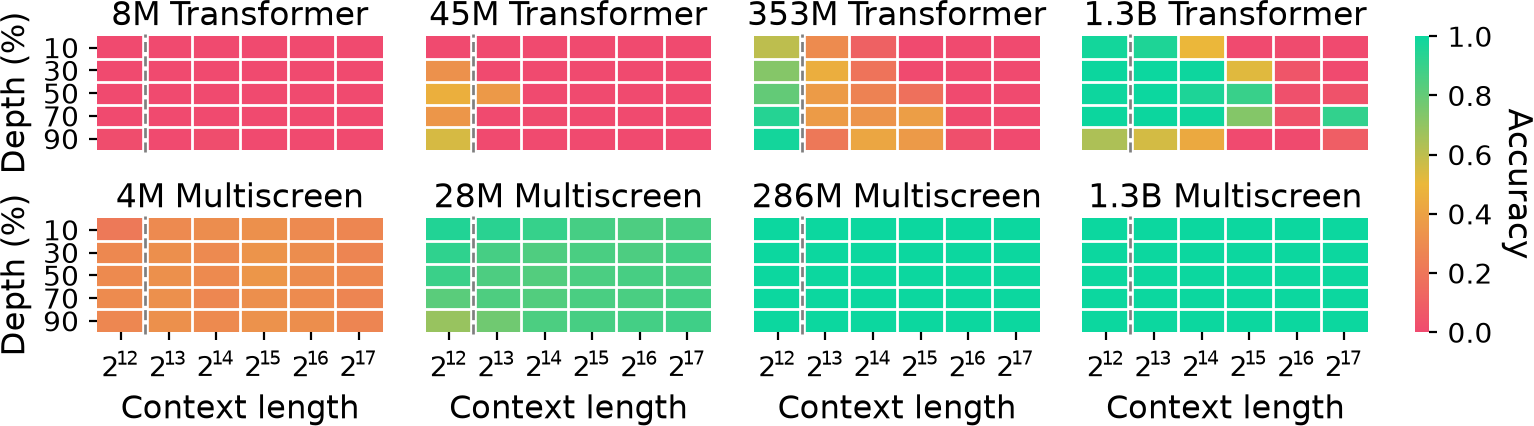}
\caption{
Passkey retrieval accuracy across model sizes for Transformer and Multiscreen.
The setup follows that of ABCDigits, with identical context lengths, target depths, and evaluation protocol.
Top row: Transformer baselines.
Bottom row: Multiscreen models using our default inference rule.
For Transformer, the YaRN scaling factor is set to the evaluation context length divided by the training context length.
Multiscreen is more robust to increasing context length than Transformer, with the 286M and 1.3B Multiscreen models achieving $100\%$ accuracy across the grid.
}
\label{fig:passkey}
\end{figure}

\paragraph{Results.}
The results are shown in \cref{fig:passkey}.
We observe the same qualitative trend as in ABCDigits: Multiscreen is substantially more robust to increasing context length than Transformer, and its retrieval accuracy improves strongly with model size.
Under our default inference rule, the 286M and 1.3B Multiscreen models achieve $100\%$ retrieval accuracy at every evaluated context length and target depth.

In contrast, Transformer performance degrades rapidly as context length increases.
Even with context-matched YaRN scaling, Transformer performance deteriorates sharply and can collapse at longer contexts.
As in ABCDigits, Multiscreen outperforms the corresponding Transformer baseline at every scale even at the training context length, despite using fewer parameters.
These results show that the improved retrieval ability of Multiscreen is not specific to the ABCDigits construction.

\clearpage
\section{Controlled Attention-Mechanism Comparison}\label[appendix]{apx:attention-comparison}

We compare screening with a broad set of attention mechanisms using the architecture and training configuration of the 353M Transformer, subject to the common document-masking exception described below.
Each model is trained for $2^{38}$ tokens using the same learning rate and remaining training hyperparameters as the 353M Transformer.
Each variant is trained once with a single random seed.
In the main comparison, only the attention mechanism is replaced; model dimensions, including the query, key, and value dimensions, and initialization otherwise match the 353M Transformer.
Throughout this section, $L_{\mathrm{eval}}$ denotes the evaluation context length, while $L_{\mathrm{train}}=2^{12}$ denotes the pretraining context length.

\paragraph{Document masking.}
We disable document masking for every model in this comparison because the publicly available fused kernels used for several variants, including Entmax, Sigmoid, and Forgetting Attention, do not support document-level masking.
This common setting ensures that every variant is trained with the same visibility across documents within a packed sequence.
The standard 353M Transformer trained with document masking is reported separately in \cref{fig:lcp-small,fig:abcd} and exhibits the same qualitative length-extrapolation failure as the Softmax control in this comparison.

\paragraph{Fused implementations.}
We use the publicly available fused implementations for Forgetting Attention,\footnote{\url{https://github.com/zhixuan-lin/forgetting-transformer}} Sigmoid attention through TritonSigmoid version 0.1.2~\citep{sadashivaiah2026better},\footnote{\url{https://github.com/MSDLLCpapers/triton-sigmoid}} and Entmax ($\alpha=1.5$) through AdaSplash version 0.2.2~\citep{goncalves2026adasplash}.\footnote{\url{https://github.com/deep-spin/adasplash}}
For Forgetting Attention, we enable adaptive computation pruning with \texttt{adaptive\_threshold="auto"}~\citep{lin2025adaptive}.

\paragraph{Compared mechanisms.}
The full Screening variant consists only of the screening unit: unit-length normalization of queries, keys, and values, MiPE, Trim, Softmask, and TanhNorm.
It does not include the output gate or any other modification from the gated screening tile, and all components outside attention remain identical to those of the Transformer backbone.
The Screening (no window) variant removes the learned screening window $w$ and its dependent components, omitting MiPE and Softmask while retaining unit-length normalization of queries, keys, and values, Trim, and TanhNorm, and uses ordinary RoPE.
The Screening (RoPE) variant retains the learned window and Softmask while replacing MiPE with ordinary RoPE.
The exact configurations are summarized in \cref{tab:attention-comparison}.

The parameter counts differ by at most 0.12\% across the comparison, keeping model capacity closely matched.
Softmax$_1$ adds one to the Softmax denominator, giving $\operatorname{softmax}_1(z)_i=\exp(z_i)/(1+\sum_j\exp(z_j))$ and allowing the total attention weight to approach zero~\citep{miller2023attention}.
In the table, YaRN denotes the full method, which combines NTK-by-parts RoPE-frequency interpolation with pre-softmax attention-logit scaling, whereas YaRN$^\ast$ retains the RoPE-frequency interpolation but omits the attention-logit scaling~\citep{peng2024yarn}.
Throughout this comparison, $s$ denotes the common scaling factor used to parameterize both YaRN positional scaling and method-specific length corrections.
The method-specific corrections use $L_{\mathrm{train}}$ as their base length, so context-matched scaling with $s=L_{\mathrm{eval}}/L_{\mathrm{train}}$ corresponds to the evaluation length $L_{\mathrm{eval}}$.
A dash indicates that no inference-time scaling is applied.
The ``method-specific'' designation indicates that a variant additionally modifies its own length-dependent term.

For Sigmoid attention, the base logit bias is $b=-\log L_{\mathrm{train}}$~\citep{ramapuram2025theory}.
At scaling factor $s$, we replace it with $b_s=-\log(sL_{\mathrm{train}})=b-\log s$, shifting every attention logit by an additional $-\log s$.
Following \citet{hua2022transformer}, the unscaled $\mathrm{ReLU}^2$ variants compute attention weights as $A=\operatorname{ReLU}^2(QK^\top/L_{\mathrm{train}}+B_{\mathrm{rel}})$, where $B_{\mathrm{rel}}$ denotes the relative-position bias used in both variants.
For each layer and head, we construct this bias from learned query-side and key-side vectors $\bm{q}_{\mathrm{rel}},\bm{k}_{\mathrm{rel}}\in\mathbb{R}^{128}$, matching the original formulation.
Its entries are
\begin{equation}
(B_{\mathrm{rel}})_{ij}
=
\operatorname{RoPE}(\bm{q}_{\mathrm{rel}},i)
\operatorname{RoPE}(\bm{k}_{\mathrm{rel}},j)^\top,
\end{equation}
where the two vectors are rotated at the query and key positions, respectively.
These vectors account for the additional parameters of the $\mathrm{ReLU}^2$ variants in \cref{tab:attention-comparison}.
For scaling factor $s$, we instead use $A_s=\operatorname{ReLU}^2((QK^\top/L_{\mathrm{train}}+B_{\mathrm{rel}})/s)$, scaling the complete pre-activation, including the relative-position bias, by $1/s$.
The original GAU formulation normalizes by the sequence length but does not specify an additional adjustment for evaluation beyond the training context; we use the above rule as its method-specific extrapolation treatment.

For Forgetting Attention, ALiBi, the sliding-window baseline, and full Screening, we apply no inference-time scaling.
Following its original specification, Forgetting Attention uses no positional encoding~\citep{lin2025forgetting}.
We classify both ALiBi and Forgetting Attention as full-context at the level of their underlying relevance rules because their distance-dependent attenuation does not impose a hard support boundary.
Forgetting Attention's adaptive computation pruning can omit sufficiently small contributions computationally; we treat these implementation-level zeros as an efficiency approximation rather than a formal property of the underlying relevance rule.

\begin{table}[t]
\centering
\caption{
Configurations in the controlled attention-mechanism comparison.
All listed variants use the same query, key, and value dimensions.
The Context column describes formal support, so mechanisms with distance-dependent attenuation but no hard support boundary are classified as Full.
Parameter counts are reported in millions and rounded to two decimal places.
}
\label{tab:attention-comparison}
\begin{tabular}{lcccr}
\toprule
Variant & Pos. enc. & Context & Scaling & \# Params (M) \\
\midrule
Softmax & RoPE & Full & YaRN & $353.45$ \\
Softmax$_1$ & RoPE & Full & YaRN & $353.45$ \\
SSMax & RoPE & Full & YaRN$^\ast$ & $353.45$ \\
Entmax ($\alpha=1.5$) & RoPE & Full & YaRN$^\ast$ & $353.45$ \\
Sigmoid & RoPE & Full & YaRN$^\ast$ + method-specific & $353.45$ \\
$\mathrm{ReLU}^2$ & RoPE & Full & YaRN$^\ast$ + method-specific & $353.55$ \\
Sliding window & RoPE & 2048 & -- & $353.45$ \\
ALiBi & ALiBi & Full & -- & $353.45$ \\
Forgetting & NoPE & Full & -- & $353.85$ \\
Screening (no window) & RoPE & Full & YaRN$^\ast$ & $353.45$ \\
Screening (RoPE) & RoPE & Learned $w$ & YaRN$^\ast$ & $353.45$ \\
Screening & MiPE & Learned $w$ & -- & $353.45$ \\
\bottomrule
\end{tabular}
\end{table}

\paragraph{Additional GAU-style parameterization.}
We also evaluate $\mathrm{ReLU}^2$ with the original GAU-style query/key parameterization, in which learned elementwise scales and shifts are applied to a shared projected representation~\citep{hua2022transformer}.
Doubling its query/key dimension approximately matches the parameters of the separate $W_Q$ and $W_K$ projections, yielding a 353.75M-parameter variant.
Because this variant underperforms the separate-projection $\mathrm{ReLU}^2$ variant on both evaluations, we report the stronger and more tightly controlled separate-projection result in \cref{fig:attention-comparison}.

\paragraph{Additional sliding-window sizes.}
We also evaluate sliding-window attention with fixed window sizes of 512 and 1024 tokens.
Because these narrower windows further restrict accessible context without changing the qualitative conclusion, the main comparison reports the 2048-token variant as the most favorable sliding-window control for long-range retrieval.

\paragraph{Properties of the relevance rules.}
\Cref{tab:attention-properties} summarizes the rule-level properties and the cross-length scaling applied to each evaluated mechanism.
``Variable total mass'' means that the relevance values are not constrained to sum to one, and ``Pairwise independent'' means that the relevance assigned to one query--key pair does not depend on the other available keys.
``Exact rejection'' requires that an accessible key can receive exactly zero relevance based on its content, excluding zeros imposed only by causal or fixed-window masks, while ``Empty selection'' requires that all accessible keys can simultaneously receive exactly zero relevance.

``No explicit length term'' means that the relevance rule contains no explicit sequence-length term, whereas ``No inference scaling'' means that the evaluated mechanism is used across context lengths without a context-length-dependent adjustment at inference.
Although inference-time scaling can preserve boundedness and pairwise independence at any fixed factor, it changes the calibration of relevance values across context lengths.
A check mark indicates that a property holds, while a dash indicates that it does not.

We call a relevance mechanism absolute across context lengths when it assigns each accessible query--key pair a relevance value independently on a bounded, length-invariant scale, permits exact rejection and empty selection, does not constrain the total relevance mass to one, and requires no inference-time scaling.
Under this definition, the listed properties are requirements for absolute relevance across context lengths, but satisfying them does not by itself guarantee better empirical performance.
For screening, at fixed query and key representations, adding other keys does not mechanically renormalize existing relevance values, and zero has the explicit meaning of rejection, making the resulting relevance signal easier to inspect and interpret.
We do not directly evaluate this potential interpretability benefit in the present work.

\begin{table}[t]
\centering
\caption{
Properties relevant to absolute relevance across context lengths for the mechanisms evaluated in the controlled comparison.
``Bounded'' indicates that each relevance value has a finite upper bound.
Structural zeros introduced by causal or fixed-window masks are excluded from ``Exact rejection.''
}
\label{tab:attention-properties}
\begin{tabular}{lccccccc}
\toprule
Variant & \rotatebox{90}{Variable total mass} & \rotatebox{90}{Pairwise independent} & \rotatebox{90}{Bounded} & \rotatebox{90}{Exact rejection} & \rotatebox{90}{Empty selection} & \rotatebox{90}{No explicit length term} & \rotatebox{90}{No inference scaling} \\
\midrule
Softmax & -- & -- & $\checkmark$ & -- & -- & $\checkmark$ & -- \\
Softmax$_1$ & $\checkmark$ & -- & $\checkmark$ & -- & -- & $\checkmark$ & -- \\
SSMax & -- & -- & $\checkmark$ & -- & -- & -- & -- \\
Entmax ($\alpha=1.5$) & -- & -- & $\checkmark$ & $\checkmark$ & -- & $\checkmark$ & -- \\
Sigmoid & $\checkmark$ & $\checkmark$ & $\checkmark$ & -- & -- & -- & -- \\
$\mathrm{ReLU}^2$ & $\checkmark$ & $\checkmark$ & -- & $\checkmark$ & $\checkmark$ & -- & -- \\
Sliding window & -- & -- & $\checkmark$ & -- & -- & $\checkmark$ & $\checkmark$ \\
ALiBi & -- & -- & $\checkmark$ & -- & -- & $\checkmark$ & $\checkmark$ \\
Forgetting & -- & -- & $\checkmark$ & -- & -- & $\checkmark$ & $\checkmark$ \\
Screening (no window) & $\checkmark$ & $\checkmark$ & $\checkmark$ & $\checkmark$ & $\checkmark$ & $\checkmark$ & -- \\
Screening (RoPE) & $\checkmark$ & $\checkmark$ & $\checkmark$ & $\checkmark$ & $\checkmark$ & $\checkmark$ & -- \\
Screening & $\checkmark$ & $\checkmark$ & $\checkmark$ & $\checkmark$ & $\checkmark$ & $\checkmark$ & $\checkmark$ \\
\bottomrule
\end{tabular}
\end{table}

Softmax$_1$ illustrates why variable total mass and empty selection are distinct: its total mass can approach zero but remains strictly positive for finite logits.
Likewise, Sigmoid attention has variable total mass and pairwise-independent weights, but its strictly positive outputs preclude exact rejection and empty selection.
Entmax can assign exact zeros to individual keys, but its unit-sum constraint prevents empty selection.
$\mathrm{ReLU}^2$ permits exact rejection and empty selection but has unbounded relevance values and an explicit sequence-length term.
Screening combines variable total mass, pairwise independence, bounded relevance, exact rejection, and empty selection without an explicit sequence-length term or inference-time scaling.

\paragraph{Evaluation protocol.}
We evaluate each model on long-context perplexity and ABCDigits retrieval using the protocols described in \cref{apx:lcp} and \cref{apx:abcd}, respectively.
The datasets, sample counts, and all other evaluation settings exactly match the main evaluations.
For variants that use inference-time scaling in the ABCDigits evaluation, we set the scaling factor to $L_{\mathrm{eval}}/L_{\mathrm{train}}$.
For example, evaluation at $L_{\mathrm{eval}}=2^{15}$ uses a scaling factor of $8$.
The dashed vertical lines in all panels mark the pretraining context length, $L_{\mathrm{train}}=2^{12}$.

Screening is the only evaluated mechanism that maintains both low long-context perplexity and robust retrieval across all evaluated context lengths and target depths without inference-time scaling (\cref{fig:attention-comparison}).
As shown in the top row, ALiBi, Forgetting Attention, and sliding-window attention maintain relatively stable perplexity at long contexts.
However, the bottom row shows that all three lose retrieval accuracy as context length increases: the distance-dependent attenuation in ALiBi and Forgetting Attention suppresses distant information, while sliding-window attention cannot access targets outside its window.
Thus, stable long-context perplexity alone does not establish access to distant information.
Softmax and the remaining global-attention alternatives likewise lose retrieval accuracy rapidly beyond the training context despite their respective extrapolation treatments.

Even at scaling factor $1$, Screening (no window) avoids the abrupt perplexity degradation beyond the training context seen in the other full-context RoPE-based mechanisms.
With context-matched scaling, it also compares favorably with most alternatives on ABCDigits beyond the training context, but its long-context perplexity behavior is the more distinctive result.
Because Screening (no window) and the failing global mechanisms all use RoPE, this contrast shows that RoPE alone cannot explain their abrupt degradation beyond the training context.

Screening (RoPE) substantially improves both evaluations over Screening (no window), isolating the benefit of the learned $w$-dependent Softmask construction while holding RoPE fixed.
Replacing RoPE with MiPE while retaining Softmask produces a striking further gain: Full Screening achieves perfect ABCDigits retrieval across every evaluated context length and target depth.
Because this replacement is the only difference from Screening (RoPE), the comparison isolates MiPE's additional contribution.
Together, these results isolate the long-context behavior to the screening unit itself rather than the output gate or the wider Multiscreen architecture.

\clearpage
\section{The Role of TanhNorm}\label[appendix]{apx:tanhnorm}

TanhNorm softly bounds the norm of the screening aggregation while preserving its direction, which may be especially important because screening relevance values are not constrained to sum to one.

To isolate this role, we compare the standard 286M Multiscreen configuration with a matched variant trained without TanhNorm while holding all remaining model, training, and evaluation settings fixed.
Because TanhNorm has no trainable parameters, removing it leaves the parameter count unchanged.
All results in this section use checkpoints at $2^{36}$ training tokens rather than the $2^{38}$-token checkpoints used in the main experiments.
The TanhNorm baseline uses the same three independently trained 286M Multiscreen runs as the model-scaling experiments in \cref{sec:scaling}, and we average their results under both inference-window treatments.
The variant without TanhNorm is evaluated using a single independently trained run.

We evaluate both models under two inference-window treatments.
The default rule converts any learned screening window exceeding the maximum sequence length seen during training to $w=\infty$ once after training and then keeps every window fixed across evaluation lengths.
The as-learned condition instead retains every learned window value unchanged.
We use the long-context perplexity and ABCDigits protocols described in \cref{apx:lcp,apx:abcd}, respectively.

\begin{figure}[ht]
\centering
\includegraphics[width=0.65\linewidth]{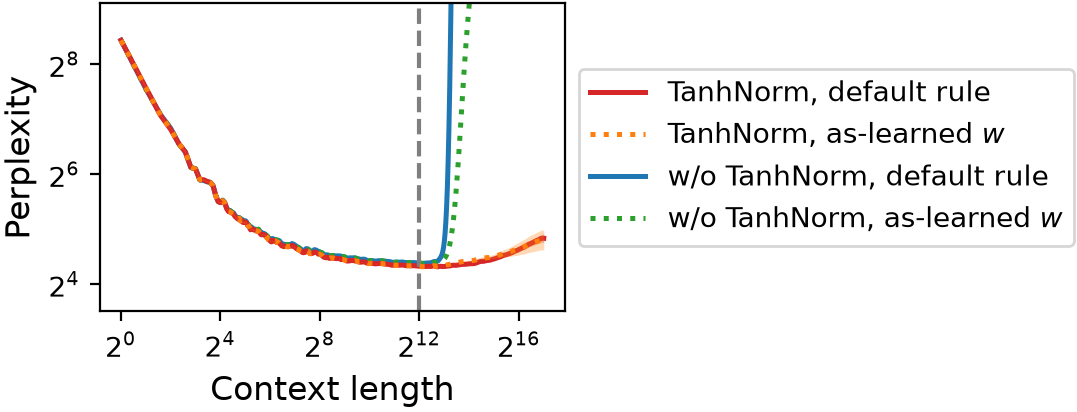}
\caption{
Long-context perplexity for 286M Multiscreen with and without TanhNorm under the default inference rule and with as-learned $w$.
For the TanhNorm baseline, curves show the mean across three independently trained runs, and shaded regions indicate one standard deviation; the variant without TanhNorm uses a single run.
The gray dashed line marks the pretraining context length ($2^{12}$).
}
\label{fig:tanhnorm-lcp}
\end{figure}

As shown in \cref{fig:tanhnorm-lcp}, the two conditions with TanhNorm nearly coincide and maintain low perplexity across the evaluated context lengths.
Without TanhNorm, both conditions remain stable beyond the pretraining context length rather than failing immediately at its boundary, but their perplexity eventually increases sharply.
The default rule fails earlier than the as-learned condition, showing that finite learned windows can delay this instability but do not eliminate it.
Thus, TanhNorm substantially extends the range of stable length extrapolation under both window treatments and is particularly important when broad-window units are converted to $w=\infty$.

\begin{figure}[ht]
\centering
\includegraphics[width=0.55\linewidth]{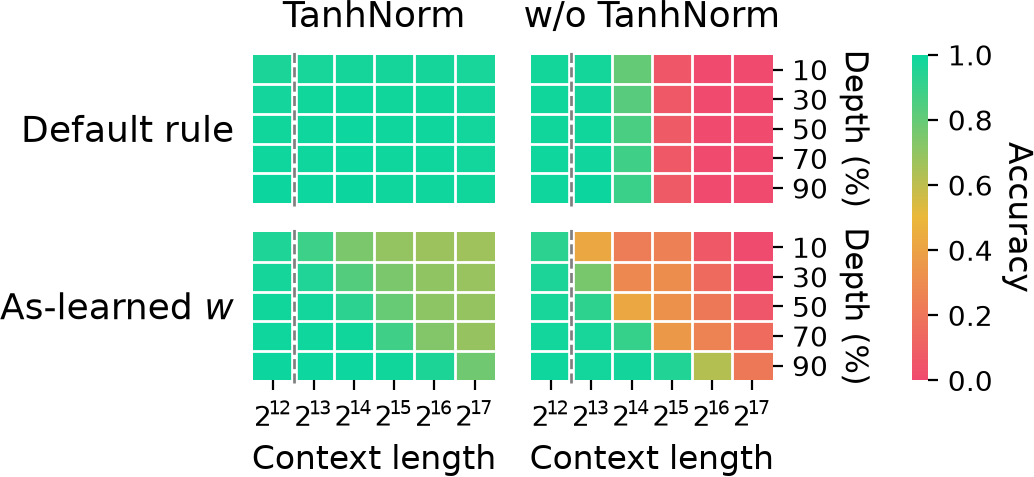}
\caption{
ABCDigits retrieval accuracy for the same 286M Multiscreen variants and inference-window treatments as in \cref{fig:tanhnorm-lcp}.
Results with TanhNorm are averaged over three independently trained runs, whereas results without TanhNorm use a single run.
The default rule applies the post-training conversion to $w=\infty$, whereas the as-learned condition leaves every learned window unchanged.
The gray dashed lines mark the pretraining context length ($2^{12}$).
}
\label{fig:tanhnorm-abcd}
\end{figure}

The retrieval results in \cref{fig:tanhnorm-abcd} reveal distinct failure modes for the two window treatments.
With TanhNorm, the default rule maintains near-perfect accuracy across the evaluated context lengths and target depths, whereas retaining as-learned windows introduces a gradual distance-dependent decline at longer contexts.
Without TanhNorm, the default rule remains effective beyond the pretraining context length but then collapses across target depths once the context becomes sufficiently long.
Retaining as-learned windows avoids this simultaneous collapse, but retrieval instead becomes increasingly local: targets nearer the query remain accessible longer than distant targets.
Even without TanhNorm, screening retains substantial length-extrapolation ability in both perplexity and retrieval.
TanhNorm extends this robustness, stabilizing long-context aggregation and preserving global retrieval when the default inference rule converts learned windows beyond the training context length to $w=\infty$.

\clearpage
\section{The Role of MiPE}\label[appendix]{apx:mipe}

MiPE couples positional rotation to each unit's learned screening window and becomes inactive for sufficiently broad windows.
To isolate its role, we compare the standard 286M Multiscreen configuration with variants that replace MiPE with RoPE or NoPE while retaining the learned window and Softmask.
These replacements introduce no trainable parameters, so all variants have the same parameter count, and all remaining model, training, and evaluation settings are fixed.

All results in this section use checkpoints at $2^{36}$ training tokens.
The MiPE baseline uses the same three independently trained 286M Multiscreen runs as the model-scaling experiments in \cref{sec:scaling}, whereas each positional-encoding ablation uses a single independently trained run.
We evaluate all variants using Multiscreen's default inference rule for $w$ and the long-context perplexity and ABCDigits protocols in \cref{apx:lcp,apx:abcd}.

\begin{figure}[ht]
\centering
\vspace{3mm}
\includegraphics[width=0.6\linewidth]{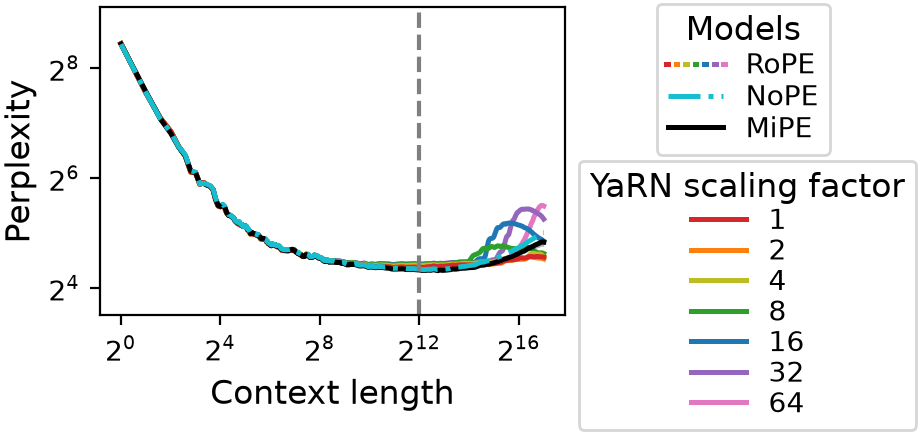}
\caption{
Long-context perplexity for 286M Multiscreen using MiPE, NoPE, or RoPE.
The MiPE curve shows the mean across three independently trained runs, and its shaded region, largely obscured by the mean curve, indicates one standard deviation; the NoPE and RoPE curves each use a single run.
MiPE and NoPE use no inference-time positional scaling; the colored RoPE curves use the indicated fixed YaRN scaling factors.
The gray dashed line marks the pretraining context length ($2^{12}$).
}
\label{fig:mipe-lcp}
\end{figure}

As shown in \cref{fig:mipe-lcp}, replacing MiPE with RoPE or NoPE changes long-context perplexity only modestly.
MiPE and NoPE maintain low perplexity without inference-time positional scaling, while the RoPE variant remains comparable under smaller YaRN scaling factors but worsens under larger factors at the longest contexts.
Thus, long-context perplexity alone provides little indication of the large retrieval differences observed below.

\begin{figure}[ht]
\centering
\includegraphics[width=0.9\linewidth]{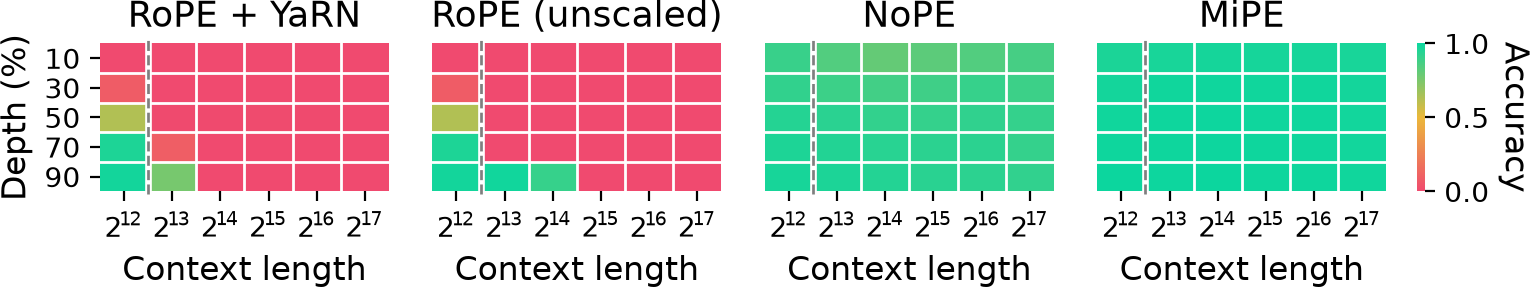}
\caption{
ABCDigits retrieval accuracy for the same 286M Multiscreen positional-encoding variants.
The first panel applies context-matched YaRN scaling to RoPE, whereas the second uses unscaled RoPE; MiPE and NoPE require no inference-time positional scaling.
The MiPE results are averaged over three independently trained runs, whereas the NoPE and RoPE results each use a single run.
The gray dashed lines mark the pretraining context length ($2^{12}$).
}
\label{fig:mipe-abcd}
\end{figure}

Retrieval sharply separates the positional-encoding variants despite their similar perplexity behavior (\cref{fig:mipe-abcd}).
NoPE retains strong retrieval, showing that the core long-context retrieval ability does not arise from positional encoding alone.
Nevertheless, MiPE outperforms NoPE in all 30 evaluated context-length--depth combinations, increasing mean accuracy across the grid from $90.71\%$ to $98.76\%$.
After averaging over target depths at each context length, MiPE maintains $98.22\%$--$99.16\%$ accuracy, whereas NoPE ranges from $89.44\%$ to $93.80\%$.
In contrast, the RoPE variant loses most retrieval accuracy beyond the pretraining context length both with context-matched YaRN and when unscaled.

These results show that MiPE does more than avoid the failure of conventional positional rotations: it supplies positional structure that consistently improves retrieval over NoPE while preserving scaling-free length extrapolation.
Together with the strong Screening (no window) result in \cref{fig:attention-comparison}, the RoPE ablation should not be interpreted as showing that RoPE inherently causes length-extrapolation failure; rather, it indicates that RoPE is poorly matched to the learned-window Softmask construction in this setting.

\clearpage
\section{The Role of the Output Gate}\label[appendix]{apx:gate}

The output gate modulates the screening output before its projection back to the model dimension.
To assess its role, we compare the standard 286M Multiscreen configuration with a closely capacity-matched variant without the gate.

Removing the gate reduces the parameter count, so we increase both the number of heads and the number of layers from 32 to 38 in the no-gate control.
The resulting models contain 286.35M and 288.05M parameters, respectively, making the no-gate control approximately $0.60\%$ larger than the standard model.
All other training and evaluation settings are unchanged, and both models are evaluated at $2^{36}$ training tokens.
The standard Multiscreen results average the same three independently trained runs used in the model-scaling experiments in \cref{sec:scaling}, whereas the no-gate control uses a single independently trained run.

\begin{figure}[ht]
\centering
\includegraphics[width=0.35\linewidth]{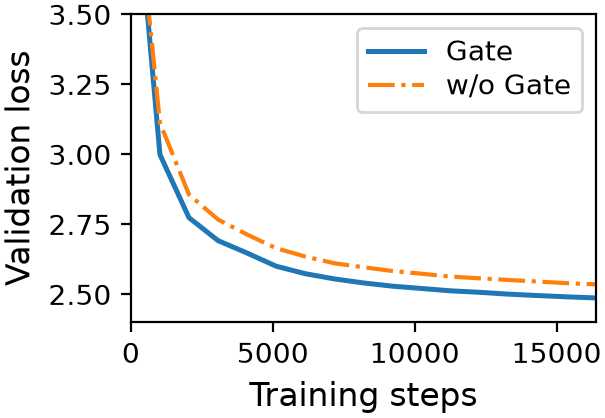}
\caption{
Validation loss over $2^{16}$ training steps ($2^{36}$ tokens) for standard Multiscreen and the closely capacity-matched variant without the output gate.
The Multiscreen curve averages three independently trained runs, and its shaded region, largely obscured by the mean curve, indicates one standard deviation; the no-gate curve uses a single run.
}
\label{fig:gate-valloss}
\end{figure}
\begin{figure}[ht]
\centering
\includegraphics[width=0.35\linewidth]{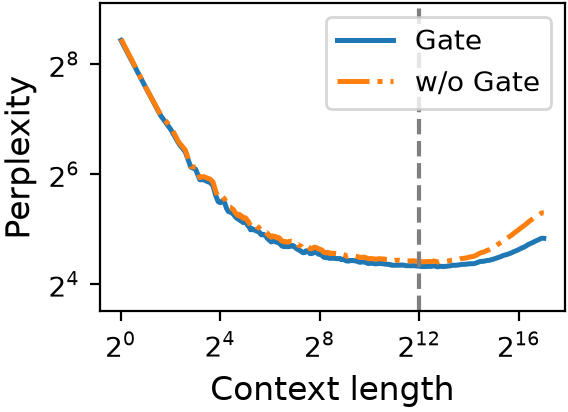}
\caption{
Long-context perplexity for standard Multiscreen and the closely capacity-matched variant without the output gate.
The standard Multiscreen curve shows the mean across three independently trained runs, and its shaded region, largely obscured by the mean curve, indicates one standard deviation; the no-gate curve uses a single run.
The gray dashed line marks the pretraining context length ($2^{12}$).
}
\label{fig:gate-lcp}
\end{figure}
\begin{figure}[ht]
\centering
\includegraphics[width=0.5\linewidth]{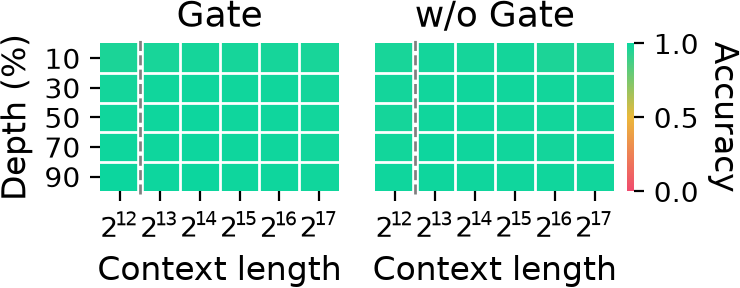}
\caption{
ABCDigits retrieval accuracy for standard Multiscreen and the closely capacity-matched variant without the output gate.
Results for standard Multiscreen are averaged over three independently trained runs, whereas the no-gate results use a single run.
The gray dashed lines mark the pretraining context length ($2^{12}$).
}
\label{fig:gate-abcd}
\end{figure}

As shown in \cref{fig:gate-valloss}, standard Multiscreen reaches consistently lower validation loss throughout training.
At $2^{36}$ training tokens, its validation loss is $2.4871 \pm 0.0005$, compared with $2.5356$ for the no-gate control.
Because the no-gate control is slightly larger, this advantage cannot be attributed to a larger parameter count in the gated model.
The output gate therefore provides a consistent improvement in overall language-modeling quality.

The no-gate control exhibits consistently higher perplexity, in line with its validation-loss disadvantage, while retaining the same qualitative length-extrapolation behavior as standard Multiscreen (\cref{fig:gate-lcp}).
Both variants also maintain near-perfect ABCDigits accuracy across the evaluated context lengths and target depths, with little discernible difference between them (\cref{fig:gate-abcd}).
Together, these results indicate that the output gate primarily improves language-modeling quality rather than providing the core mechanism for long-context extrapolation or retrieval.

\clearpage
\section{TTFT Measurement Protocol}\label[appendix]{apx:ttft}

All model-side TTFT measurements were performed on a single NVIDIA GH200 GPU with batch size 1, causal masking, and mixed precision with bfloat16 matrix multiplications.
We used the same Transformer and Multiscreen kernel implementations as in training.
Each trained model was measured 100 times, and we report the mean TTFT.

\clearpage
\section{Existing Assets and Licenses}\label[appendix]{apx:licenses}

We summarize the existing datasets, tokenizers, and software packages used in this work, together with their licenses and terms of use.

\paragraph{Datasets.}
We use SlimPajama~\citep{cerebras2023slimpajama} for pretraining.
SlimPajama is a cleaned and deduplicated dataset derived from RedPajama~\citep{weber2024redpajama}, and is released under the Apache License 2.0 by Cerebras.\footnote{\url{https://www.cerebras.ai/blog/slimpajama-a-627b-token-cleaned-and-deduplicated-version-of-redpajama}}
RedPajama itself aggregates multiple data sources with source-specific licenses and terms of use.

For long-context perplexity evaluation, we use PG-19~\citep{rae2020compressive}.
The PG-19 repository is released under the Apache License 2.0\footnote{\url{https://github.com/google-deepmind/pg19}}, and the benchmark consists of books extracted from Project Gutenberg that were published before 1919.
Project Gutenberg states that the vast majority of its eBooks are in the public domain in the United States, while users outside the United States should verify the copyright status under their local laws.\footnote{\url{https://www.gutenberg.org/policy/permission.html}}

\paragraph{Tokenization.}
We use the GPT-2 tokenizer~\citep{radford2019language}.
The GPT-2 code and tokenizer assets are released by OpenAI under a modified MIT license.\footnote{\url{https://github.com/openai/gpt-2}}

\paragraph{Software.}
Our implementation uses PyTorch~\citep{paszke2019pytorch} for neural-network training and inference.
PyTorch is distributed under a BSD-style license.\footnote{\url{https://github.com/pytorch/pytorch}}
For Transformer baselines, we use PyTorch's \texttt{torch.nn.{\allowbreak}attention.{\allowbreak}varlen.{\allowbreak}varlen\_{\allowbreak}attn} implementation with its built-in FlashAttention-2 backend~\citep{dao2024flashattention2}.
For Multiscreen kernels, we use Triton~\citep{tillet2019triton}, which is released under the MIT license.\footnote{\url{https://github.com/triton-lang/triton}}
We access the downstream benchmarks through version 0.4.12 of the Language Model Evaluation Harness~\citep{sutawika2026lmeval}, which is released under the MIT License,\footnote{\url{https://github.com/EleutherAI/lm-evaluation-harness}} and cite the original benchmark sources in \cref{apx:downstream}.
The controlled attention-mechanism comparison uses the forgetting-transformer implementation\footnote{\url{https://github.com/zhixuan-lin/forgetting-transformer}} and TritonSigmoid version 0.1.2,\footnote{\url{https://github.com/MSDLLCpapers/triton-sigmoid}} both released under the MIT License, and AdaSplash version 0.2.2, released under the BSD 3-Clause License.\footnote{\url{https://github.com/deep-spin/adasplash}}

\end{document}